\documentclass[final,5p,times,twocolumn,authoryear]{elsarticle}

\usepackage[switch]{lineno}

\usepackage{amssymb}
\usepackage{lipsum}
\usepackage{multirow}
\usepackage{algorithmic}

\usepackage{textcomp}
\usepackage{stfloats}
\usepackage{url}
\usepackage{verbatim}
\usepackage{cite}
\usepackage{amsmath,amssymb,amsfonts}
\usepackage{algorithmic}
\usepackage{mathtools}
\usepackage{times}
\usepackage{soul}
\usepackage{multirow}
\usepackage{lipsum}
\usepackage{physics}
\usepackage[export]{adjustbox}

\usepackage{graphicx, caption, subcaption}

\usepackage{booktabs,multirow,array,caption}
\usepackage{times}
\usepackage[table]{xcolor}
\newenvironment{blackText}{\color{black}}{}

\journal{Neural Networks}

\begin{document}
\begin{frontmatter}

\title{Effects of Mixed Sample Data Augmentation on Interpretability of Neural Networks}

\author[a]{Soyoun Won}
\author[a]{Sung-Ho Bae}
\author[a]{Seong Tae Kim}
\affiliation[a]{organization={Department of Computer Science and Engineering, Kyung Hee University},
            city={Yongin-si},
            postcode={17104}, 
            state={Gyeonggi-do},
            country={South Korea}}

\begin{abstract}

  Mixed sample data augmentation strategies are actively used when training deep neural networks (DNNs). Recent studies suggest that they are effective at various tasks. However, the impact of mixed sample data augmentation on model interpretability has not been widely studied. In this paper, we explore the relationship between model interpretability and mixed sample data augmentation, specifically in terms of feature attribution maps. To this end, we introduce a new metric that allows a comparison of model interpretability while minimizing the impact of occlusion robustness of the model. Experimental results show that several mixed sample data augmentation decreases the interpretability of the model and \textcolor{black}{label mixing} during data augmentation plays a significant role in this effect. This new finding suggests it is important to carefully adopt the mixed sample data augmentation method, particularly in applications where attribution map-based interpretability is important.
\end{abstract}


\end{frontmatter}

\section{Introduction}
\label{sec:intro}

Mixed sample data augmentation strategies overlap or cut-and-paste more than one input sample to create augmented data \citep{uddin2020saliencyMix, Yun2019CutMix, zhang2017mixup, yang2022recursivemix}. These methods are proven to be effective at improving the generalization ability of deep neural networks (DNNs). Models trained with mixed sample data augmentation achieve higher performance than the vanilla counterpart on various tasks \citep{Yun2019CutMix, cao2022survey}. 
The benefits of mixed sample data augmentation strategies are widely studied in various fields. 
However, the effect of mixed sample data augmentation on interpretability is still under shadow. In other words, the following important question remains unanswered. How do cutting and mixing input images, and rearranging labels affect model interpretability? 

The black-box nature of DNNs makes it difficult to understand their decision-making process, which could be critical issues in mission-critical applications. To use mixed sample data augmentation in mission-critical domains, understanding the effects of mixed sample data augmentation on model interpretability becomes important. A wide range of efforts has been made to improve the interpretability of DNNs~\citep{selvaraju2017gradCAM, Schulz2020IBA, Zhang2021inputIBA, chen2019protoPnet, nauta2021protoTree, bau2020dissection, khakzar2021neural, nishimura2023extraction, mehta2022explainable}. Feature attribution is one of the most widely adopted methods in the explanation of neural networks. It assigns an importance score to input features in a post-hoc manner (e.g., creating a heatmap where each pixel value represents the contribution of the corresponding pixel to the model prediction). 

To address the aforementioned research question, we explore how mixed sample data augmentation affects post-hoc explanations of models. When input data is mixed at train time and hence alters the model's decision-making process, how will the attribution method react to these changes? In this paper, we shed light on the subject and reveal that mixed sample data augmentation is not an all-round player in the lens of post-hoc explanations.  

Firstly, we point out the limitations of localization metrics that previous mixed sample data augmentation studies reported to improve \citep{Yun2019CutMix} in measuring the interpretability of models. Then, we re-measure localization-based model interpretability by using an energy-based metric. Secondly, we evaluate the faithfulness of attribution methods to the models trained with various mixed sample data augmentation strategies and show that when the models are trained with mixed inputs, the connection between the model and the saliency map becomes loose. 
In particular, we highlight the shortcomings of the existing feature ablation-based evaluation method \citep{Samek2017deletion} in comparing the interpretability of various models. We then introduce a novel feature ablation-based evaluation approach to address this issue. Feature ablation (e.g., deletion and insertion) has been used to evaluate different \textit{feature attribution methods} on the same model. However, we aim to assess various \textit{models} on a general feature attribution method. We show that the score from the previous feature ablation method represents not only the interpretability of the produced explanation but also the occlusion robustness of the model, and thus the whole scores are biased towards occlusion robustness (i.e., models with higher occlusion robustness tend to achieve higher feature ablation score). 
\begin{blackText}
To address this issue, we propose a new idea that reduces the effect of occlusion robustness in the feature ablation process. 
Our proposed metric provides a way to a fair comparison of the interpretability. 
Through the proposed evaluation framework, it is possible to evaluate the interpretability of models trained with different kinds of mixed sample data augmentation. 
\end{blackText}

Experimental results show that the models trained with modern mixed sample data augmentation strategies produce the degraded quality of the saliency maps when visualized using post-hoc explanation methods. We further observe that \textcolor{black}{label mixing} to produce augmented (mixed) input plays an important role in the degradation of interpretability.

The main contribution of the paper is summarized as follows:
\begin{itemize}
\item We evaluate the interpretability of models trained with mixed sample data augmentation strategies. 
\item We propose a new inter-model faithfulness comparison metric: Inter-Model Deletion.
\item \textcolor{black}{We find that models trained with mixed sample data augmentation archive lower interpretability and label mixing plays an important role in this effect. }
\end{itemize} 


\section{Related Work}
\label{sec:related}

\subsection{Mixed Sample Data Augmentation}
\textbf{Mixup} \citep{zhang2017mixup} combines two randomly chosen images into one input image by linear interpolation. The mix ratio is drawn from the beta distribution. Labels are determined by the proportion of each class in a new augmented image. Mixup is head of various mix-based augmentation strategies \citep{mai2021metaMixup, verma2019manifoldMixup, choi2022tokenmixup, yin2021batchMixup}. \textbf{CutMix} \citep{Yun2019CutMix} is inspired by Cutout \citep{devries2017cutout} and Mixup~\citep{zhang2017mixup}. It drops small square regions of an image like Cutout and fills the regions with another randomly chosen image, like Mixup. 
The label is modified as a proportion of the augmented image. \textbf{SaliencyMix} \citep{uddin2020saliencyMix} builds upon CutMix, but the fundamental difference is that filled regions are selected carefully so that “salient” regions are to be mixed. Salient regions are selected using the algorithm proposed in~\citep{Montabone2010}. The new label is determined the same way as CutMix.
\textbf{RecursiveMix} \citep{yang2022recursivemix} also builds upon CutMix, but it preserves cutmix images from the previous batch and pastes them in the image in the current batch to repeat the procedure, which results in a mixed image from more than two samples. The final label is modified by the proportion of each image contained. \textbf{PixMix} \citep{hendrycks2022pixmix} is similar to Mixup, but instead of mixing two samples in the training set, PixMix mixes one sample with the patterned images such as fractals and feature visualizations a random number of times. The label is preserved as the original label. 
\textbf{TokenMixup} \citep{choi2022tokenmixup} is a mixed sample data augmentation strategy specifically devised for vision transformers. It selects what sample and token to be mixed to maximize the saliency of the augmented tokens. Horizontal TokenMixup (HTM) utilizes two different samples to create the augmented sample while Vertical TokenMixup (VTM) mixes one sample in a token level by using the tokens from previous layers.

\subsection{Bonus Effect of Mixed Sample Data Augmentation}
~\citet{Thulasidasan2019mixupCalibration} studied the overlooked effect of Mixup on calibration. They found that DNNs trained with Mixup output softmax scores that are closer to the actual likelihood than the vanilla DNNs. ~\citet{zhang2022mixupCalibration} provided theoretical proof for this discovery. Also, ~\citet{chun2020calibration} observed better calibration at Cutout and CutMix-trained models. Meanwhile,~\citet{wangpitfall} found Mixup has actually a negative impact on post-hoc calibration. Our study is in line with these studies in that we investigate what mixed sample data augmentation strategies can unexpectedly bring. 

\subsection{Feature Attribution}
Feature attribution is a post-hoc explanation method that assigns importance scores to input features that contribute to the model’s output. It is a useful tool for understanding the inner decision-making process of DNNs. 
Recent attribution methods aim to improve the interpretability of a model by using more elaborate ways to extract important features \citep{Zhang2021inputIBA, kolek2022cartoon}. 
Studies can be largely categorized as activation, perturbation, and gradient-based methods. \textbf{Activation-based} method uses feature activation maps to assign importance scores to each pixel in the input space~\citep{zhou2016learning,selvaraju2017gradCAM,Wang2020scoreCAM}. \textbf{Perturbation-based} methods calculate weights by observing the change in the model score against input/feature perturbation~\citep{fong2017interpretable,fong2019understanding,Schulz2020IBA,kim2021robust}. \textbf{Gradient-based} methods utilize a gradient that computes the change of the model score to the change of input space
~\citep{springenberg2014striving,smilkov2017smoothgrad,sundararajan2017axiomatic,srinivas2019full}. 

In this study, we evaluate attribution maps from different models trained with various mixed sample data augmentation strategies. We exploit GradCAM \citep{selvaraju2017gradCAM} from activation-based approaches, IBA \citep{Schulz2020IBA} from perturbation-based approaches, and Smooth Gradient (SmoothGrad) \citep{smilkov2017smoothgrad}, Integrated Gradient (IntGrad) \citep{sundararajan2017axiomatic}, and Guided Backpropagation (GBP) \citep{springenberg2014striving} from gradient-based approaches. 

\subsection{Evaluation of Feature Attribution}

Evaluating feature attribution methods is not a simple task because there is no existing ground truth \citep{Rao2022}. 
Several studies introduced functional-grounded evaluation metrics \citep{Samek2017deletion,Adebayo2018sanityCheck,hooker2019benchmark,khakzar2022explanations,Rao2022,rong2022consistent}. 
Localization is one of the widely used automatic metrics~\citep{fong2017interpretable,Wang2020scoreCAM,Zhang2021inputIBA}. The purpose of localization is to quantify how well the attribution map highlights the point of interest. Other approaches include feature ablation which degrades the input image using the importance score that the attribution map provides and observes the change in the model's output \citep{Samek2017deletion,hooker2019benchmark,rong2022consistent}. 
Building upon \citet{Samek2017deletion}, ~\citet{hooker2019benchmark} proposed RemOve And Retrain (ROAR) which trains new models from the degraded images to compare different attribution methods under artifact-controlled conditions. 
~\citet{rong2022consistent} introduced RemOve And Debias (ROAD) to reduce the computational time for retraining models in ROAR. However, these approaches do not align well with our study because our focus is not on comparing the interpretability of different attribution methods, but rather on comparing the interpretability of different \textit{models}.
Therefore, we propose a new evaluation scheme building upon \citet{Samek2017deletion} to assess the interpretability of models trained with different mixed sample augmentation strategies.

\section{Effect of Mixed Sample Data Augmentation on Model Interpretability}
\label{experiments}

\subsection{Experimental Design}
\label{experiment_design}
``Interpretability'' is often said, but for a lot of works, defined differently. Some studies aim to understand models by analyzing the ``concepts'' that the model detects. These studies examine what kind of concepts the model recognizes to interpret its behavior \citep{bau2017dissection,kim2018interpretability}. Other lines of work define interpretability using attribution maps, and there are two main approaches for measuring it: intrinsic and extrinsic. Extrinsic interpretability often involves humans, such as determining whether human understanding aligns with the explanation provided by the attribution map \citep{Kim2022hive}, whether humans can intervene in the model \citep{koh2020concept}, or whether humans and the model can work together \citep{taesiri2022humanaiteam}. Intrinsic evaluation, on the other hand, involves the automatic (computational) assessment of the attribution map without human involvement. We narrow our focus on interpretability in terms of the intrinsic view of attribution maps.

To evaluate models that are trained with different mixed sample data augmentation methods, we designed two experiments. 
\textbf{Localization} experiment aims to quantify the visual alignment between attribution maps and ground truth bounding box. This serves as a measure of find-grainedness and it is a metric for visual/human interpretability \citep{Zhang2021inputIBA}. 
\textbf{Inter-Model Deletion} aims to evaluate faithfulness. It measures how well the attribution map represents the model itself by ablating features based on attribution scores. We devise a new metric to reduce the effect of occlusion robustness of the model.
Both experiments are designed to capture false positives and false negatives in attribution maps in that a low score is measured when the attribution map assigns high importance to less important features (i.e., false positive) and when it assigns low importance to more important features (i.e., false negative).

\begin{table}[t]
\centering
\caption{Classification top-1 accuracy (\%) of the baseline and models trained with mixed sample data augmentation strategies. MobileNetV2 is trained on CIFAR-10, and ResNet-50 is trained on ImageNet.}
\label{tab:performance}
\begin{adjustbox}{width=0.33\textwidth}
    \begin{tabular}{lcc}
        \noalign{\smallskip}\noalign{\smallskip}\hline\hline
        Methods & MobileNetV2  & ResNet-50 \\
        \hline
        Baseline & 92.52 &76.32  \\ 
        Mixup & 93.20 &77.42 \\
        CutMix & 93.13 &78.60 \\ 
        SaliencyMix & 93.06 &78.74\\ 
        RecursiveMix & 94.38&79.20  \\ 
        PixMix & 92.86 & 77.40 \\
        \hline
        \hline
    \end{tabular}
\end{adjustbox}
\end{table}
\subsection{Experimental Setup}
\label{experimental_setup}
For CIFAR-10 \citep{krizhevsky2009cifar}, we trained each model on the base structure of MobileNetV2 \citep{sandler2018mobilenetv2}. Models are trained with a batch size of 128, for 200 epochs. A stochastic gradient descent (SGD) optimizer is used with weight decay 5e-4, Nesterov momentum 0.9. The initial learning rate is set to 0.1 and later decayed by the factor of 0.2 after 50, 120, and 160 epochs. We used a single RTX 3090 GPU for training each model. We followed the original hyperparameters set by the proposed papers as much as possible. For Mixup, CutMix, and SaliencyMix, the mixing ratio is drawn from Beta distribution $\lambda \sim \text{Beta}(\alpha, \alpha)$. We trained Mixup, CutMix, and SaliencyMix with $\alpha$ of 1.0. RecursiveMix determines the mixing ratio from the uniform distribution $U[0, \alpha]$. We set $\alpha$ as 0.5 by following \citep{yang2022recursivemix}. For PixMix, we use mixing weights from Beta distribution $\beta$ = 3 and maximum mixing number of times $k$ = 4. For RecursiveMix, we set the upper bound of uniform distribution $\alpha$ = 0.5 and loss weight $\omega$ = 0.1.

For ImageNet \citep{russakovsky2015imagenet}, we conduct experiments on ResNet-50 \citep{he2016deep}. we use model weights released by the original authors of the augmentation methods. ResNet-50 models are trained with a batch size of 256, and for 300 epochs with weight decay - the initial learning rate is set to 0.1 and later decayed by the factor of 0.1 at three epoch steps (75, 150, and 225). All of them including the baseline ResNet-50 are trained with traditional data augmentation strategies such as flipping, cropping, and resizing. 

Classification performance is provided in Table \ref{tab:performance}. The performance of models trained with mixed sample data augmentation surpasses the baseline for all probed cases across model architecture and dataset. 

For IBA, we utilized Per-Sample Bottleneck and set $\beta$ = 10. For SmoothGrad, we iterated over 50 times to generate noisy samples and set the standard deviation of Gaussian perturbations to 0.15. 

\subsection{Localization}
\label{section:localization}

\begin{table*}[t]
\centering
\renewcommand{\arraystretch}{1}
\caption{EnergyPG and EHR score of six models, trained with different augmentation strategies on ImageNet. The models are evaluated by five feature attribution methods, respectively. The best and second-best results are shown in bold, and underlined, respectively. The average of five attribution methods are reported. The standard errors are also presented.}
\label{tab:localization}
\begin{tabular}{cc|ccccc|c}
\noalign{\smallskip}\noalign{\smallskip}\hline\hline

Metric & Methods & GradCAM & IBA & SmoothGrad & IntGrad & GBP & Average 
\\
\hline
\multirow{6}{*}{EnergyPG}
         &Baseline & \underline{0.549}\begin{scriptsize}$\pm$0.004\end{scriptsize} & \textbf{0.637}\begin{scriptsize}$\pm$0.004\end{scriptsize} & 0.494\begin{scriptsize}$\pm$0.003\end{scriptsize} & \textbf{0.465}\begin{scriptsize}$\pm$0.003\end{scriptsize} & \underline{0.651}\begin{scriptsize}$\pm$0.004\end{scriptsize} &
         \textbf{0.559}\begin{scriptsize}$\pm$0.002\end{scriptsize} 
         \\ 
         &Mixup & 0.530\begin{scriptsize}$\pm$0.004\end{scriptsize} & \underline{0.608}\begin{scriptsize}$\pm$0.004\end{scriptsize} & 0.485\begin{scriptsize}$\pm$0.003\end{scriptsize} & 0.420\begin{scriptsize}$\pm$0.003\end{scriptsize} & 0.403\begin{scriptsize}$\pm$0.004\end{scriptsize} &
         0.489\begin{scriptsize}$\pm$0.002\end{scriptsize}
         \\
         &CutMix & 0.521\begin{scriptsize}$\pm$0.004\end{scriptsize} & 0.560\begin{scriptsize}$\pm$0.004\end{scriptsize} & 0.493\begin{scriptsize}$\pm$0.004\end{scriptsize} & 0.441\begin{scriptsize}$\pm$0.003\end{scriptsize} & 0.485\begin{scriptsize}$\pm$0.004\end{scriptsize} &
         0.500\begin{scriptsize}$\pm$0.002\end{scriptsize}\\
         &SaliencyMix & 0.520\begin{scriptsize}$\pm$0.004\end{scriptsize} & 0.562\begin{scriptsize}$\pm$0.004\end{scriptsize} & 0.498\begin{scriptsize}$\pm$0.003\end{scriptsize} & 0.440\begin{scriptsize}$\pm$0.003\end{scriptsize} & 0.362\begin{scriptsize}$\pm$0.004\end{scriptsize} &
         0.476\begin{scriptsize}$\pm$0.002\end{scriptsize}\\
         &RecursiveMix & \textbf{0.569}\begin{scriptsize}$\pm$0.004\end{scriptsize} & 0.596\begin{scriptsize}$\pm$0.004\end{scriptsize} & \textbf{0.508}\begin{scriptsize}$\pm$0.003\end{scriptsize} & 0.447\begin{scriptsize}$\pm$0.003\end{scriptsize} & 0.532\begin{scriptsize}$\pm$0.004\end{scriptsize} &
         \underline{0.530}\begin{scriptsize}$\pm$0.002\end{scriptsize}\\
         &PixMix & 0.462\begin{scriptsize}$\pm$0.003\end{scriptsize} & 0.531\begin{scriptsize}$\pm$0.004\end{scriptsize} & \underline{0.506}\begin{scriptsize}$\pm$0.003\end{scriptsize} & \underline{0.464}\begin{scriptsize}$\pm$0.003\end{scriptsize} & \textbf{0.658}\begin{scriptsize}$\pm$0.004\end{scriptsize} &
         0.524\begin{scriptsize}$\pm$0.002\end{scriptsize}\\
\hline
\multirow{6}{*}{EHR}
        &Baseline & \underline{0.484}\begin{scriptsize}$\pm$0.003\end{scriptsize} 
        &\textbf{0.449}\begin{scriptsize}$\pm$0.003\end{scriptsize} 
        & 0.255\begin{scriptsize}$\pm$0.002\end{scriptsize} 
        & 0.187\begin{scriptsize}$\pm$0.002\end{scriptsize} 
        &\textbf{0.254}\begin{scriptsize}$\pm$0.002\end{scriptsize}
        &\textbf{0.326}\begin{scriptsize}$\pm$0.002\end{scriptsize}
        \\
        &Mixup & 0.477\begin{scriptsize}$\pm$0.003\end{scriptsize} & 0.404\begin{scriptsize}$\pm$0.003\end{scriptsize} & 0.272\begin{scriptsize}$\pm$0.002\end{scriptsize} & \textbf{0.272}\begin{scriptsize}$\pm$0.002\end{scriptsize} & 0.114\begin{scriptsize}$\pm$0.002\end{scriptsize} &
         0.308\begin{scriptsize}$\pm$0.002\end{scriptsize}\\
        &CutMix & 0.474\begin{scriptsize}$\pm$0.003\end{scriptsize} & 0.417\begin{scriptsize}$\pm$0.003\end{scriptsize} & 0.273\begin{scriptsize}$\pm$0.002\end{scriptsize} & 0.174\begin{scriptsize}$\pm$0.002\end{scriptsize} & 0.167\begin{scriptsize}$\pm$0.002\end{scriptsize} &
         0.301\begin{scriptsize}$\pm$0.002\end{scriptsize}\\
        &SaliencyMix & 0.478\begin{scriptsize}$\pm$0.003\end{scriptsize} & 0.414\begin{scriptsize}$\pm$0.003\end{scriptsize} & \underline{0.279}\begin{scriptsize}$\pm$0.002\end{scriptsize} & 0.173\begin{scriptsize}$\pm$0.002\end{scriptsize} & 0.079\begin{scriptsize}$\pm$0.002\end{scriptsize} &
         0.285\begin{scriptsize}$\pm$0.002\end{scriptsize}\\
        &RecursiveMix & \textbf{0.507}\begin{scriptsize}$\pm$0.003\end{scriptsize} & \underline{0.425}\begin{scriptsize}$\pm$0.003\end{scriptsize} & \textbf{0.286}\begin{scriptsize}$\pm$0.002\end{scriptsize} & 0.179\begin{scriptsize}$\pm$0.002\end{scriptsize} & 0.178\begin{scriptsize}$\pm$0.003\end{scriptsize} &
         0.315\begin{scriptsize}$\pm$0.002\end{scriptsize}\\
        &PixMix & 0.475\begin{scriptsize}$\pm$0.003\end{scriptsize} & 0.407\begin{scriptsize}$\pm$0.003\end{scriptsize} & 0.266\begin{scriptsize}$\pm$0.002\end{scriptsize} & \underline{0.188}\begin{scriptsize}$\pm$0.002\end{scriptsize} & \textbf{0.254}\begin{scriptsize}$\pm$0.002\end{scriptsize} &
        \underline{0.318}\begin{scriptsize}$\pm$0.001\end{scriptsize}\\
\hline
\hline
\end{tabular}
\end{table*}

\begin{figure}
\begin{center}
\includegraphics[scale=0.2]{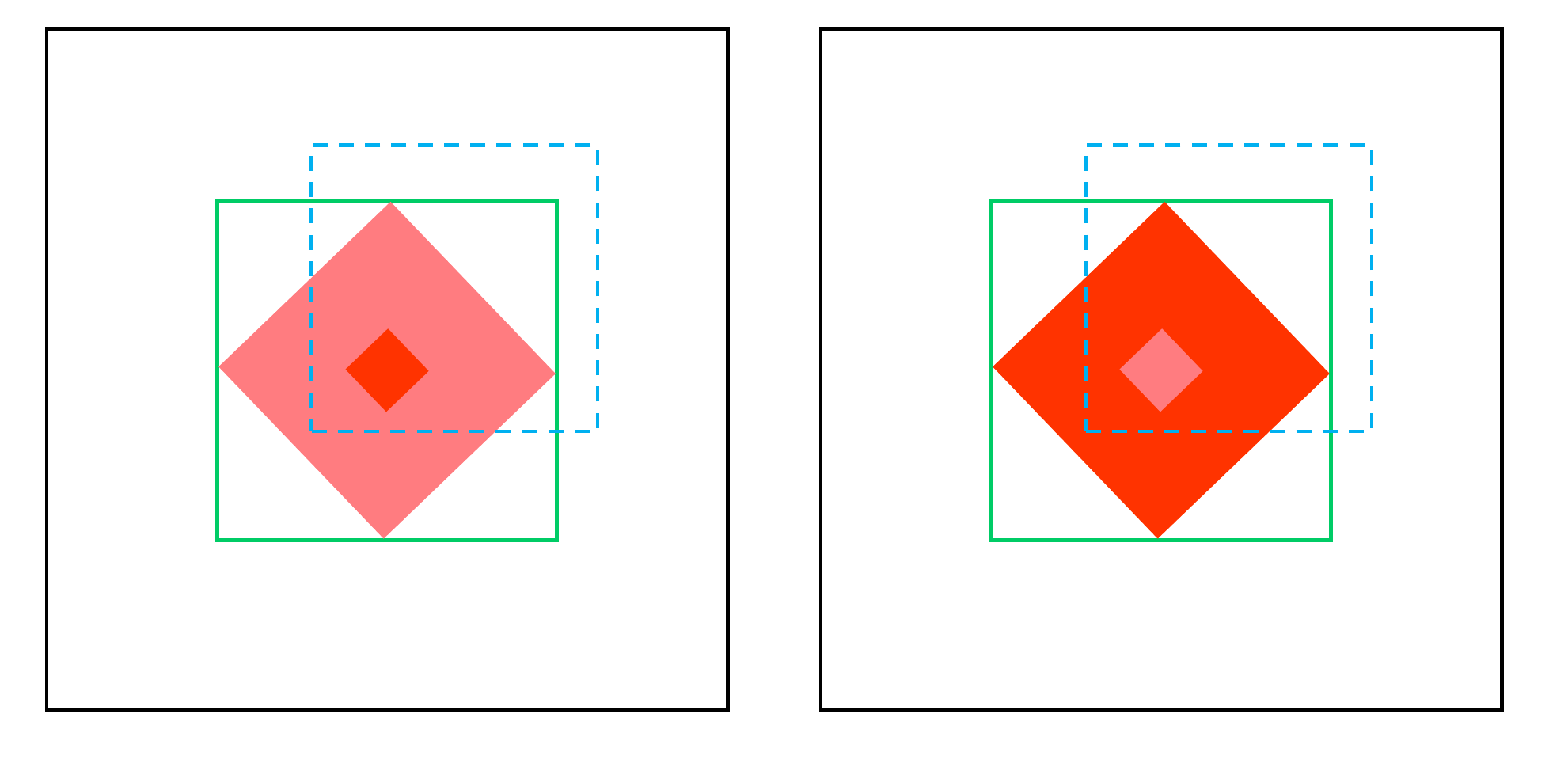}
\end{center}
\caption{Two different heatmaps after clipping where their shape and size are the same. The green box denotes the estimated bounding box from the attribution maps' silhouette. The black box with a dashed line represents the ground truth bounding box. Left: higher attribution at the center, lower attribution at the edge. Right: higher attribution at the edge, lower attribution at the center. Nonetheless, the same bounding box is estimated for evaluation in WSOL. }
\vspace{-1em}
\label{fig:wsol}
\end{figure}

In previous mixed sample data augmentation studies \citep{Yun2019CutMix,uddin2020saliencyMix}, weakly supervised object localization (WSOL) has been used to interpret the model to focus on the object. 
WSOL task requires a binarized attribution map based on the pre-defined threshold to create a mask. Then, the smallest box that contains the whole mask region is estimated. Finally, the Intersection over union (IoU) between the estimated box and ground truth object bounding box is measured. 

This approach is a good measure for WSOL tasks. However, it is unsuitable for evaluating feature attribution maps due to the following reasons. Firstly, the synthetically estimated bounding box adds bias to the original heatmap. Secondly, WSOL task does not consider each pixel value (i.e., the importance of the feature) within the heatmap.  Thirdly, it is sensitive to the threshold parameter.
For example, consider two attribution maps that have the same silhouette (Figure  \ref{fig:wsol}). One attributes high values on the object and low values on the near-object background, and the other attributes low values on the object and high values on the near-object background. These two heatmaps clearly tell different things, but the same bounding box will be estimated in WSOL. Therefore, we consider the Energy-based Pointing Game (EnergyPG) \citep{Wang2020scoreCAM} and Effective Hit Ratio (EHR) \citep{Zhang2021inputIBA} to alleviate the aforementioned problems and fairly evaluate localization from the perspective of interpretability. Note that these two metrics aim to measure attribution map alignment with human-annotated bounding boxes \citep{Wang2020scoreCAM}. while attribution outside of the bounding box may affect interpretability in another definition, our paper studies the effect of various mixed sample data augmentation on localization in terms of map-box alignment.

\textbf{Energy-based Pointing Game (EnergyPG)} ~\citet{Wang2020scoreCAM} introduced EnergyPG to measure the proportion of energy (i.e., attribution) contained in the ground-truth object bounding box over the whole energy of the attribution map. EnergyPG is a synthetic box-free and threshold-free metric that makes use of every pixel value in the calculation. $L^c$ denotes the attribution map for class $c$. 

\begin{equation}
Score_{EPG} = \frac{\sum{L^c_{(i,j)\in bbox}}}{\sum{L^c_{(i,j)\in bbox}} + \sum{L^c_{(i,j)\notin bbox}}}.
\end{equation}

\textbf{Effective Heat Ratio (EHR)} ~\citet{Zhang2021inputIBA} proposed EHR which calculates the energy in the ground-truth bounding box divided by the total number of pixels above multiple quantiles of thresholds, which are values between 0.0 to 1.0. Then, a plot representing EHR score at each threshold is obtained. Finally, the area under the curve (AUC) is computed. The thresholding step gives calibrated attribution maps that control the dispersion. EHR aims to calculate the localization considering the distribution of attribution maps after applying the calibration. Let $S^{c,\lambda}_{(i,j)} = (L^c_{(i,j)} > \lambda)$ where $\lambda$ denotes a threshold. Then, EHR score is calculated as below. 

\begin{equation}
\text{Score}_{EHR} = \sum_{\lambda}\left[\frac{\sum L^c_{(i,j)\in \text{bbox}}}{\sum S^{c,\lambda}_{(i,j)\in \text{bbox}} + \sum S^{c,\lambda}_{(i,j)\notin \text{bbox}}}\right].
\end{equation}

Similar to \citet{Rao2022,bohle2021convolutional,Nguyen2021effectiveness}, we randomly select 2,000 random samples from the ImageNet validation set that are correctly classified with model scores higher than 0.6. 
Also, we restrict our samples with bounding boxes covering less than 50\% and more than 10\% of the original image for discrimination. 

Experimental results are given in Table \ref{tab:localization}. To our surprise, some data augmentation strategies that showed better performance on WSOL showed the opposite results in our localization experiments. Attribution maps calculated from baseline ResNet-50 scored the highest localization on average despite the lowest classification performance. The localization scores are low on the baseline for Smoothgrad, where the score difference between the six models is marginal. On average, The deterioration of the model that scored the best (i.e., Baseline on EnergyPG and EHR) and the second-best model (i.e., RecursiveMix on EnergyPG and PixMix on EHR) is statistically significant ($p < 0.001$ and $p = 0.0016$ on EnergyPG and EHR, respectively). 

\subsection{Qualitative Analysis}
\subsubsection{ImageNet Dataset}
\begin{figure*}[t]
\centering
\includegraphics[scale=0.75]{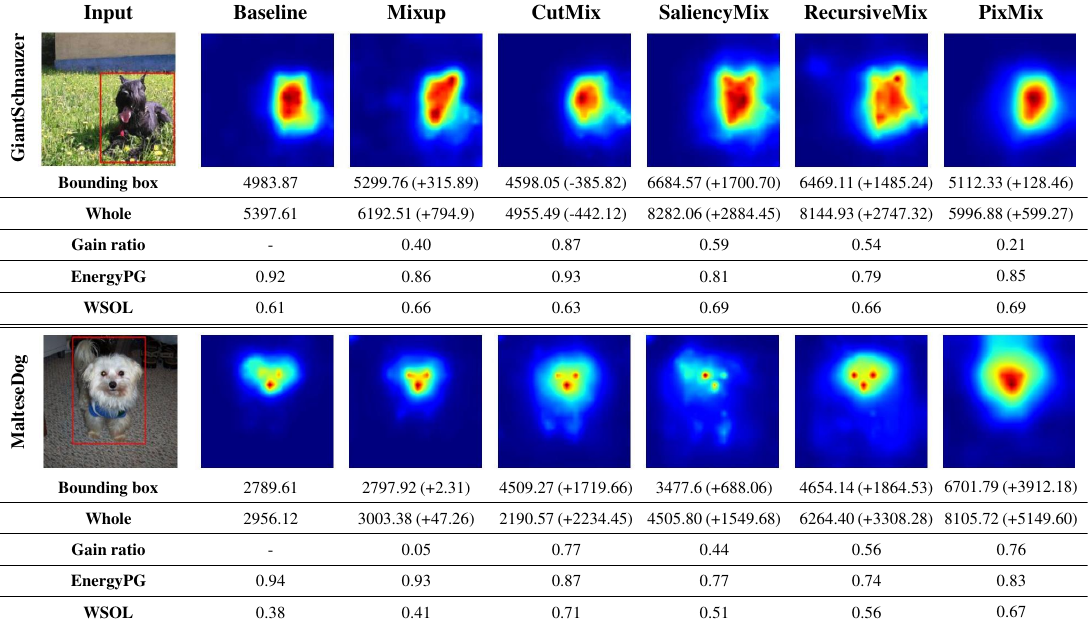}
\caption{Qualitative comparison of baseline and the models trained with different mixed sample data augmentation strategies via IBA visualization. Ground truth annotations are depicted as red boxes. \emph{Bounding box} and \emph{whole} represent the energy inside of the bounding box and the whole image, respectively. \emph{Gain ratio} denotes the amount of increased energy inside the bounding box divided by the amount of increased energy of the total attribution map. The threshold of WSOL is set to 0.15 as in~\citet{Yun2019CutMix}.}

\label{fig:quali}
\end{figure*}

Qualitative analysis in Figure \ref{fig:quali} shows wider regions of the feature attribution map from models trained with augmented data. However, this does not directly lead to better localization performance. While they assign more importance to broader areas, attributions inside the bounding box (denoted by \emph{Bounding box}) do not increase as much as the total attributions from the entire heatmap (denoted by \emph{Whole}). In other words, models trained with the augmentation method attribute more energy to the object, but, at the same time, they assign higher importance scores to non-relevant regions. This leads to relatively small energy in the bounding box and large energy outside the bounding box. Figure \ref{fig:quali} also depicts the limitations of WSOL. Intersection over union calculation of WSOL favors attribution maps with broad silhouettes, ignoring each pixel value. 

\subsubsection{Benchmarking Attribution Methods (BAM) Dataset}
The Benchmarking Attribution Methods (BAM) dataset \citep{yang2019bam} is a synthetic dataset where random objects are pasted into random backgrounds. More specifically, the dataset is created by pasting 10 object classes from MSCOCO \citep{lin2014mscoco} onto 10 scene classes from MiniPlaces \citep{zhou2017places}. The object classes consist of backpack, bird, dog, elephant, kite, pizza, stop sign, toilet, truck, and zebra, and the scene classes consist of bamboo forest, bedroom, bowling alley, bus interior, cockpit, corn field, laundromat, runway, ski slope, and track. 

Then, synthetically sculpted images are duplicated into two datasets - each with object and scene labels. Given the same sample, for example, a dog in a bamboo forest, object datasets label the sample as "dog" and scene datasets label it as "bamboo forest". Here, The fundamental idea is to verify relative feature importance. We do not know the ground truth (i.e., what feature the models consider important) but we do know the relative feature importance. For example, for the models that are trained with object labels, the object area would be relatively more important than the scene area.  

We train models with BAM dataset on ResNet-50 architecture initialized with pre-trained weights from Torchvision package \citep{paszke2017automatic}. Models are trained on the same image with two different labels: object and scene. Adam optimizer is used with weight decay 5e-4 and a learning rate of 0.1. The batch size is set to 128. Hyperparameters of augmentation strategies are the same as the training setting in CIFAR-10 dataset. Classification scores are shown in Table \ref{tab:bam-performance}.

\begin{table}[t]
\centering
\caption{Classification accuracy (\%) of ResNet-50 trained with mixed sample augmentation strategies on BAM.}
\label{tab:bam-performance}
\begin{adjustbox}{width=0.25\textwidth}
    \begin{tabular}{lcc}
        \noalign{\smallskip}\noalign{\smallskip}\hline\hline
        Methods & Object & Scene \\
        \hline
        Baseline & 87.49 & 94.27\\ 
        Mixup & 88.58 & 93.79\\ 
        CutMix & 89.59 & 94.67\\ 
        SaliencyMix & 89.36 & 94.17\\ 

        \hline
        \hline
    \end{tabular}
\end{adjustbox}
\end{table}

\begin{figure}[t]
\centering
\includegraphics[scale=0.37]{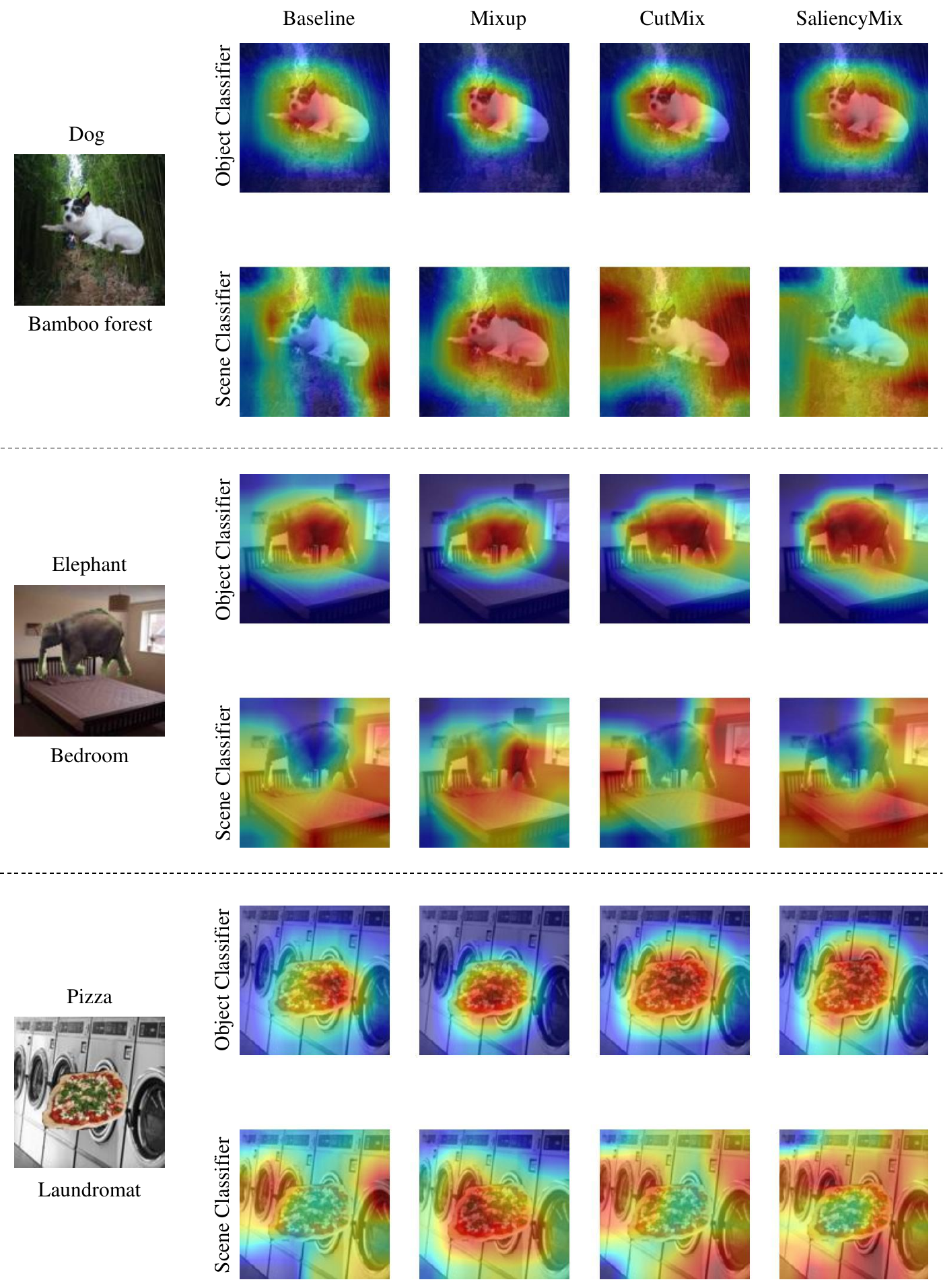}
\caption{GradCAM visualization of models trained with mixed sample data augmentation on BAM dataset. High attribution at the object region on the object classifier and low attribution at the object region on the scene classifier indicates less false positive of the attribution map. }
\label{fig:bam}
\end{figure}

Given this idea, we show qualitative examples of GradCAM visualization of models trained with mixed sample data augmentation in Figure \ref{fig:bam}. All given samples are correctly classified. Attribution maps from the object classifiers tend to localize the target (i.e. the object) well while the attribution maps from the scene classifiers do not (i.e., high attribution is assigned to object regions). Mixup produces scene attribution maps that are highly focused on the object region. Considering that the sample is correctly classified by the scene classifier, it is hard to say the resulting attribution map contains meaningful interpretation.

\subsection{Inter-Model Deletion}
\label{section:Inter-Model Deletion}

\begin{figure*}[t]
\centering
\includegraphics[scale=0.35]{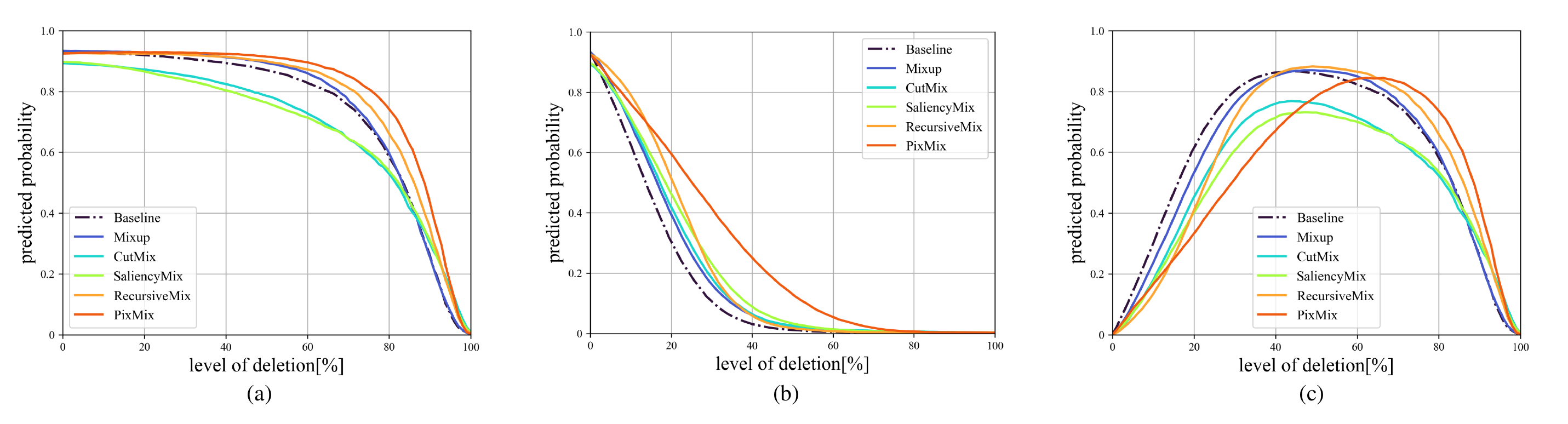}
\caption{Inter-Model Deletion curves obtained by GradCAM. The models are trained with various mixed sample augmentation methods on ImageNet. \textbf{Top}: (a) LeRF deletion curve. Faithfulness and occlusion robustness are entangled. (b) RaO deletion curve. The occlusion robustness of the models is represented. (c) Inter-Model Deletion curve. The final score is calculated by the area under the curve (c). The exact measures are provided in Table \ref{inter-deletion}. }
\label{fig:inter-del}
\end{figure*}

\captionsetup[table]{font=small, labelfont=bf}
\begin{table}
           \centering
           \captionsetup[subtable]{position = below}
          \captionsetup[table]{position=top}
           \caption{(a) Correlation between RaO and Vanilla LeRF scores of models trained with different augmentation methods. Vanilla LeRF is highly correlated with RaO, which indicates that the model's response to deletion itself (occlusion robustness) is reflected in the LeRF score along with the faithfulness of the attribution map. (b) correlation between RaO and Inter-Model Deletion scores of models trained with different augmentation methods. Inter-Model Deletion alleviates the correlation between attribution-based deletion and random deletion.}
           \label{tab:correlation}

           \begin{subtable}{0.35\linewidth}
               \centering
                    \begin{adjustbox}{width=0.83\textwidth}
                    \begin{tabular}{cc}
                        \noalign{\smallskip}\noalign{\smallskip}\hline\hline
                        Dataset & LeRF \\
                        \hline
                        CIFAR-10 & 0.512  \\ 
                        ImageNet & 0.869   \\
                        
                        \hline
                        \hline
                    \end{tabular}
                    \end{adjustbox}
               \caption{Correlation between LeRF and RaO scores of models trained with various mixed sample data augmentation strategies on CIFAR-10 and ImageNet. }
           \end{subtable}%
           \hspace*{4em}
           \begin{subtable}{0.35\linewidth}
               \centering
                    \begin{adjustbox}{width=0.92\textwidth}
                    \begin{tabular}{c|cc}
                        \noalign{\smallskip}\noalign{\smallskip}\hline\hline
                        Dataset & InterDel \\
                        \hline
                        CIFAR-10 & -0.020 \\ 
                        ImageNet & -0.192  \\
                        
                        \hline
                        \hline
                    \end{tabular}
                    \end{adjustbox}
                \caption{Correlation between Inter-Model Deletion and RaO scores of models trained with various mixed sample data augmentation strategies on CIFAR-10 and ImageNet.}
                 \label{tab:dimGMM}
           \end{subtable}
\end{table}

\begin{table*}[t]
\centering
\renewcommand{\arraystretch}{1}
\caption{Vanilla LeRF and RaO scores on CIFAR-10 and ImageNet. The average of five attribution methods on vanilla LeRF is reported. Higher is better. The best results are shown in bold. The standard errors are also presented. } 
\label{tab:vanilla_deletion}
\begin{tabular}{cc|ccccc|c|c}

\noalign{\smallskip}\noalign{\smallskip}\hline\hline
&& \multicolumn{6}{c|}{Vanilla LeRF} & RaO \\
\hline
Dataset  & Methods & GradCAM & IBA & SmoothGrad & IntGrad & GBP & \textbf{Average} & - 
\\
\hline
\multirow{6}{*}{CIFAR-10}

        & Baseline 
    & 57.74\begin{scriptsize}$\pm$0.52 \end{scriptsize}
    & 50.04\begin{scriptsize}$\pm$0.58 \end{scriptsize}
    & 62.26\begin{scriptsize}$\pm$0.48 \end{scriptsize}
    & 61.87\begin{scriptsize}$\pm$0.48 \end{scriptsize}
    & 55.08\begin{scriptsize}$\pm$0.49\end{scriptsize}
    & 57.40\begin{scriptsize}$\pm$0.23\end{scriptsize}
    &29.87\begin{scriptsize}$\pm$0.49 \end{scriptsize}

    \\
        & Mixup 
    & 39.78\begin{scriptsize}$\pm$0.41 \end{scriptsize}
    & 39.77\begin{scriptsize}$\pm$0.40 \end{scriptsize}
    & 51.33\begin{scriptsize}$\pm$0.32 \end{scriptsize}
    & 49.71\begin{scriptsize}$\pm$0.30 \end{scriptsize}
    & 41.96\begin{scriptsize}$\pm$0.36 \end{scriptsize}
    & 44.51\begin{scriptsize}$\pm$0.17 \end{scriptsize}
    &32.36\begin{scriptsize}$\pm$0.42 \end{scriptsize}

    \\
        & CutMix 
    & 50.47\begin{scriptsize}$\pm$0.46 \end{scriptsize}
    & 45.62\begin{scriptsize}$\pm$0.49\end{scriptsize}
    & 55.04\begin{scriptsize}$\pm$0.47 \end{scriptsize}
    & 57.17\begin{scriptsize}$\pm$0.44 \end{scriptsize}
    & 50.89\begin{scriptsize}$\pm$0.47  \end{scriptsize}
    & 51.84\begin{scriptsize}$\pm$0.21 \end{scriptsize}
    &36.15\begin{scriptsize}$\pm$0.48 \end{scriptsize}

    \\
    
        & SaliencyMix 
    & 53.13\begin{scriptsize}$\pm$0.36 \end{scriptsize}
    & 48.66\begin{scriptsize}$\pm$0.38 \end{scriptsize}
    & 52.81\begin{scriptsize}$\pm$0.41 \end{scriptsize}
    & 56.87\begin{scriptsize}$\pm$0.40 \end{scriptsize}
    & 50.59\begin{scriptsize}$\pm$0.42 \end{scriptsize}
    & 52.41\begin{scriptsize}$\pm$0.18 \end{scriptsize}
    &34.45\begin{scriptsize}$\pm$0.45 \end{scriptsize}

      \\ 
        & RecursiveMix 
    & 55.67\begin{scriptsize}$\pm$0.41 \end{scriptsize}
    & 49.2\begin{scriptsize}$\pm$0.48 \end{scriptsize}
    & 53.96\begin{scriptsize}$\pm$0.44 \end{scriptsize}
    & 54.71\begin{scriptsize}$\pm$0.41 \end{scriptsize}
    & 47.85\begin{scriptsize}$\pm$0.43  \end{scriptsize}
    & 52.28\begin{scriptsize}$\pm$0.20 \end{scriptsize}
    &32.80\begin{scriptsize}$\pm$0.43 \end{scriptsize}

      \\ 
        & PixMix 
    & 63.75\begin{scriptsize}$\pm$0.37 \end{scriptsize}
    & 55.9\begin{scriptsize}$\pm$0.41 \end{scriptsize}
    & 73.71\begin{scriptsize}$\pm$0.39 \end{scriptsize}
    & 67.52\begin{scriptsize}$\pm$0.40 \end{scriptsize}
    & 59.82\begin{scriptsize}$\pm$0.40  \end{scriptsize}
    & \textbf{64.14}\begin{scriptsize}$\pm$0.19 \end{scriptsize}
    & \textbf{39.88}\begin{scriptsize}$\pm$0.35 \end{scriptsize}

      \\

\hline

\multirow{6}{*}{ImageNet}

        &Baseline 
        & 73.56\begin{scriptsize}$\pm$0.37\end{scriptsize}
        & 74.76\begin{scriptsize}$\pm$0.40 \end{scriptsize}
        & 69.39\begin{scriptsize}$\pm$0.44 \end{scriptsize}
        & 60.36\begin{scriptsize}$\pm$0.47 \end{scriptsize}
        & 62.61\begin{scriptsize}$\pm$0.50\end{scriptsize}
        & 63.14\begin{scriptsize}$\pm$0.20\end{scriptsize}
        & 16.00\begin{scriptsize}$\pm$0.21\end{scriptsize}

    \\

        & Mixup 
        & 74.78\begin{scriptsize}$\pm$0.33 \end{scriptsize}
        & 76.96\begin{scriptsize}$\pm$0.36 \end{scriptsize}
        & 73.19\begin{scriptsize}$\pm$0.40 \end{scriptsize}
        & 59.31\begin{scriptsize}$\pm$0.48 \end{scriptsize}
        & 58.02\begin{scriptsize}$\pm$0.50 \end{scriptsize}
        & 68.45\begin{scriptsize}$\pm$0.20 \end{scriptsize}
        &18.69\begin{scriptsize}$\pm$0.24\end{scriptsize}

    \\ 
        & CutMix 
        & 68.39\begin{scriptsize}$\pm$0.29 \end{scriptsize}
        & 71.16\begin{scriptsize}$\pm$0.31 \end{scriptsize}
        & 67.67\begin{scriptsize}$\pm$0.34 \end{scriptsize}
        & 60.69\begin{scriptsize}$\pm$0.38 \end{scriptsize}
        & 59.04\begin{scriptsize}$\pm$0.42\end{scriptsize}
        & 65.39\begin{scriptsize}$\pm$0.16 \end{scriptsize}
        &19.23\begin{scriptsize}$\pm$0.22\end{scriptsize}

    \\ 
        & SaliencyMix 
        &67.78\begin{scriptsize}$\pm$0.29 \end{scriptsize}
        & 72.02\begin{scriptsize}$\pm$0.31 \end{scriptsize}
        & 69.84\begin{scriptsize}$\pm$0.33 \end{scriptsize}
        & 63.5\begin{scriptsize}$\pm$0.37 \end{scriptsize}
        & 58.39\begin{scriptsize}$\pm$0.40 \end{scriptsize}
        & 66.30\begin{scriptsize}$\pm$0.16 \end{scriptsize}
        &20.49\begin{scriptsize}$\pm$0.23\end{scriptsize}

    \\
        & RecursiveMix 
        &76.87\begin{scriptsize}$\pm$0.28 \end{scriptsize}
        & 79.28\begin{scriptsize}$\pm$0.29 \end{scriptsize}
        & 76.38\begin{scriptsize}$\pm$0.32 \end{scriptsize}
        & 67.48\begin{scriptsize}$\pm$0.38 \end{scriptsize}
        & 68.94\begin{scriptsize}$\pm$0.40  \end{scriptsize}
        & 73.79\begin{scriptsize}$\pm$0.16 \end{scriptsize}
        &21.16\begin{scriptsize}$\pm$0.21\end{scriptsize}

    \\

        & PixMix 
        &79.43\begin{scriptsize}$\pm$0.29 \end{scriptsize}
        & 79.39\begin{scriptsize}$\pm$0.32 \end{scriptsize}
        & 77.83\begin{scriptsize}$\pm$0.34 \end{scriptsize}
        & 69.61\begin{scriptsize}$\pm$0.4 \end{scriptsize}
        & 74.11\begin{scriptsize}$\pm$0.39  \end{scriptsize}
        & \textbf{76.08}\begin{scriptsize}$\pm$0.16 \end{scriptsize}
        & \textbf{27.22}\begin{scriptsize}$\pm$0.31\end{scriptsize}

    \\

\hline
\hline
\end{tabular}
\end{table*}


\begin{table*}[t]
\centering
\renewcommand{\arraystretch}{1}
\caption{Inter-Model Deletion results on CIFAR-10 and ImageNet. Higher is better. The models are evaluated by five feature attribution methods, respectively. The best and second-best results are shown in bold, and underlined, respectively. The average of five attribution methods are reported. The standard errors are also presented. }
\label{inter-deletion}
\begin{tabular}{cc|ccccc|c}
\noalign{\smallskip}\noalign{\smallskip}\hline\hline
Dataset  & Methods & GradCAM & IBA & SmoothGrad & IntGrad & GBP & \textbf{Average} 
\\
\hline
\multirow{6}{*}{CIFAR-10}
        &Baseline & \textbf{27.87}\begin{scriptsize}$\pm$0.47\end{scriptsize}
     & \textbf{20.17}\begin{scriptsize}$\pm$0.49\end{scriptsize}
     & \underline{32.39}\begin{scriptsize}$\pm$0.44\end{scriptsize}
     & \textbf{32.00}\begin{scriptsize}$\pm$0.38\end{scriptsize}
     & \textbf{25.21}\begin{scriptsize}$\pm$0.35\end{scriptsize}
     & \textbf{27.53}\begin{scriptsize}$\pm$0.20\end{scriptsize}
      \\ 
        & Mixup & 7.42\begin{scriptsize}$\pm$0.30\end{scriptsize}
     & 7.41\begin{scriptsize}$\pm$0.27\end{scriptsize}
     & 18.96\begin{scriptsize}$\pm$0.30\end{scriptsize}
     & 17.34\begin{scriptsize}$\pm$0.26\end{scriptsize}
     & 9.60\begin{scriptsize}$\pm$0.21\end{scriptsize}
     & 12.15\begin{scriptsize}$\pm$0.13\end{scriptsize}
      \\ 
        & CutMix & 14.32\begin{scriptsize}$\pm$0.36\end{scriptsize}
     & 9.47\begin{scriptsize}$\pm$0.36\end{scriptsize}
     & 18.89\begin{scriptsize}$\pm$0.34\end{scriptsize}
     & 21.02\begin{scriptsize}$\pm$0.30\end{scriptsize}
     & 14.73\begin{scriptsize}$\pm$0.27\end{scriptsize}
     & 15.69\begin{scriptsize}$\pm$0.15\end{scriptsize}
      \\ 
        & SaliencyMix & 18.68\begin{scriptsize}$\pm$0.40\end{scriptsize}
     & 14.21\begin{scriptsize}$\pm$0.40\end{scriptsize}
     & 18.36\begin{scriptsize}$\pm$0.41\end{scriptsize}
     & 22.43\begin{scriptsize}$\pm$0.29\end{scriptsize}
     & 16.14\begin{scriptsize}$\pm$0.27\end{scriptsize}
     & 17.96\begin{scriptsize}$\pm$0.16\end{scriptsize}
      \\ 
        & RecursiveMix & 22.87\begin{scriptsize}$\pm$0.35\end{scriptsize}
     & \underline{16.41}\begin{scriptsize}$\pm$0.36\end{scriptsize}
     & 21.17\begin{scriptsize}$\pm$0.35\end{scriptsize}
     & 21.91\begin{scriptsize}$\pm$0.30\end{scriptsize}
     & 15.05\begin{scriptsize}$\pm$0.27\end{scriptsize}
     & 19.48\begin{scriptsize}$\pm$0.15\end{scriptsize}
      \\ 
        & PixMix & \underline{23.87}\begin{scriptsize}$\pm$0.37\end{scriptsize}
     & 16.02\begin{scriptsize}$\pm$0.39\end{scriptsize}
     & \textbf{33.83}\begin{scriptsize}$\pm$0.32\end{scriptsize}
     & \underline{27.64}\begin{scriptsize}$\pm$0.28\end{scriptsize}
     & \underline{19.94}\begin{scriptsize}$\pm$0.29\end{scriptsize}
     & \underline{24.26}\begin{scriptsize}$\pm$0.16\end{scriptsize}
      \\ 

\hline

\multirow{6}{*}{ImageNet}

    &Baseline & \textbf{57.55}\begin{scriptsize}$\pm$0.33\end{scriptsize}
    & \textbf{58.75}\begin{scriptsize}$\pm$0.34\end{scriptsize}
    & 53.38\begin{scriptsize}$\pm$0.37\end{scriptsize}
    & \underline{44.35}\begin{scriptsize}$\pm$0.42\end{scriptsize}
    & 46.60\begin{scriptsize}$\pm$0.39\end{scriptsize}
    & \underline{52.13}\begin{scriptsize}$\pm$0.18\end{scriptsize}
     \\
        & Mixup & \underline{56.09}\begin{scriptsize}$\pm$0.30\end{scriptsize}
     & \underline{58.27}\begin{scriptsize}$\pm$0.31\end{scriptsize}
     & \underline{54.50}\begin{scriptsize}$\pm$0.35\end{scriptsize}
     & 40.63\begin{scriptsize}$\pm$0.45\end{scriptsize}
     & 39.33\begin{scriptsize}$\pm$0.40\end{scriptsize}
     & 49.76\begin{scriptsize}$\pm$0.18\end{scriptsize}
      \\
      
        & CutMix & 49.16\begin{scriptsize}$\pm$0.26\end{scriptsize}
     & 51.93\begin{scriptsize}$\pm$0.27\end{scriptsize}
     & 48.44\begin{scriptsize}$\pm$0.28\end{scriptsize}
     & 41.46\begin{scriptsize}$\pm$0.32\end{scriptsize}
     & 39.81\begin{scriptsize}$\pm$0.31\end{scriptsize}
     & 46.16\begin{scriptsize}$\pm$0.14\end{scriptsize}
      \\ 
        & SaliencyMix & 47.29\begin{scriptsize}$\pm$0.26\end{scriptsize}
     & 51.53\begin{scriptsize}$\pm$0.27\end{scriptsize}
     & 49.35\begin{scriptsize}$\pm$0.27\end{scriptsize}
     & 43.01\begin{scriptsize}$\pm$0.32\end{scriptsize}
     & 37.90\begin{scriptsize}$\pm$0.31\end{scriptsize}
     & 45.82\begin{scriptsize}$\pm$0.14\end{scriptsize}
      \\ 
        & RecursiveMix & 55.71\begin{scriptsize}$\pm$0.25\end{scriptsize}
     & 58.11\begin{scriptsize}$\pm$0.26\end{scriptsize}
     & \textbf{55.22}\begin{scriptsize}$\pm$0.27\end{scriptsize}
     & \textbf{46.32}\begin{scriptsize}$\pm$0.34\end{scriptsize}
     & \textbf{47.77}\begin{scriptsize}$\pm$0.31\end{scriptsize}
     & \textbf{52.63}\begin{scriptsize}$\pm$0.14\end{scriptsize}
      \\ 
        & PixMix & 52.21\begin{scriptsize}$\pm$0.30\end{scriptsize}
     & 52.17\begin{scriptsize}$\pm$0.35\end{scriptsize}
     & 50.61\begin{scriptsize}$\pm$0.31\end{scriptsize}
     & 42.39\begin{scriptsize}$\pm$0.38\end{scriptsize}
     & \underline{46.89}\begin{scriptsize}$\pm$0.31\end{scriptsize}
     & 48.85\begin{scriptsize}$\pm$0.15\end{scriptsize}
      \\ 

\hline
\end{tabular}
\end{table*}

\subsubsection{Preliminary}

In this section, we aim to evaluate \textbf{faithfulness} of attribution maps to the model - the ability to reflect whether the feature attribution correctly represents the importance of the features on the model output. The goal of feature attribution methods is to explain the underlying reason for the model decision. Therefore, it should rely on the model, rather than be plausible to humans (e.g., it should avoid merely highlighting the edge of an object regardless of the real evidence behind the model's decision \citep{Adebayo2018sanityCheck}). Faithfulness is often measured using deletion scores, which observe the model’s score change after removing small regions from the original input \citep{fong2017interpretable,Samek2017deletion,Schulz2020IBA,Zhang2021inputIBA}. The underlying idea is that more attributed regions will affect the model's decision significantly, and less attributed regions will have the opposite effect. We evaluate models trained from different augmentation strategies on deletion scores. However, the vanilla deletion metrics were originally devised to compare the interpretability of different \textit{attribution methods}. 
It is not appropriate to directly use the measure to compare the interpretability of different \textit{models} because it does not consider occlusion robustness - the characteristics of different models that are inevitably accompanied by deletion procedure.

A DNN is called robust if the network keeps its original decision even when small perturbations are added to the inputs. Occlusion is a specific form of perturbation where the perturbations block certain locations of the image \citep{guo2023occrob}. Previous works reveal that DNNs are vulnerable to occlusions \citep{eykholt2018robust,brown2017adversarial, zhu2019robustness} (i.e., DNNs fail to keep their decisions when the image is blocked by regional perturbations). Occlusion robustness varies by the model architecture and training conditions \citep{zhu2019robustness, Yun2019CutMix}. Deletion includes deleting certain regions in the input image, in other words applying occlusion perturbation to the input. Therefore, evaluating the faithfulness of explanations of different models using vanilla deletion is inevitably affected by the occlusion robustness of the model. To alleviate this issue, we propose a new deletion scheme, Inter-Model Deletion, to enable a fair comparison of the faithfulness of attribution maps between different models.

\begin{blackText}

Before describing Inter-Model Deletion, we show the change in problem definition for vanilla deletion and Inter-Model Deletion using concepts commonly employed in statistics.
In statistics, the experiments where the effect of some variable is tested in a complex environment, the experimental setting must be controlled. The dependent variable is the final outcome of the experiment, which is influenced by changes in the independent variables. Independent variables are the cause or factor being studied. Control variables are variables that affect the final outcome of the experiment, that are not the subject of studies, and thus should be kept constant.

The goal of vanilla deletion is to compare the faithfulness of attribution maps obtained using different attribution methods. In the concepts from statistics, the goal can be formulated as below where $F_{A}$ is the faithfulness of the attribution method, $A$ is the choice of attribution method, $Z$ is the control variables, and $\xi$ represents vanilla deletion. 

\begin{equation}
\begin{aligned}
F_{A} = \xi(A ; Z_{A}) \\
\end{aligned}
\end{equation}

Since the subject of analysis is different attribution methods, $Z_{A}$ can be represented by $Z_{A} = \{z_{oc}, z_{d}, z_{r}\}$ where $z_{oc}$ is the occlusion robustness of the model, $z_{d}$ is the dataset choice, and $z_{r}$ is other factors such as computational resources. In other words, the types of the model and training method which determine the occlusion robustness of the model, the dataset, and computational resources should kept frozen to measure the performance of different attribution methods using the deletion function. 

In this study, we change the subject of evaluation from attribution method $A$ to the model choice $M$. Then, the formula is reshaped as below. 

\begin{equation}
\begin{aligned}
F_{M} = \xi(M ; Z_M) \\
\end{aligned}
\end{equation}

\noindent Here, $Z_{M}$ can be listed by $Z_{M} = \{z_{oc}, z_{d}, z_{r}, z_{e}\}$ where $z_{e}$ is the attribution method choices. In this setting, where the model choice what to be examined, rather than to be controlled, $z_{oc}$ is hard to control because the different choice of models inevitably comes with different occlusion robustness.

Thus, we propose to address occlusion robustness separately. We define the occlusion robustness of the model $OC_{M}$ as below where $\zeta$ is the function that measures the occlusion robustness.

\begin{equation}
\begin{aligned}
OC_{M} = \zeta(M ; Z_M) \\
\end{aligned}
\end{equation}

\noindent Here,  $Z_{M} = \{z_{d}, z_{r}\}$

because $\zeta$ is independent of explanations being produced by the attribution method. 

In this paper, we propose to exploit $\zeta$ along with $\xi$ to control the occlusion robustness of the model for a fair comparison of the faithfulness of explanations of different models. For the vanilla faithfulness and the occlusion measure, we use vanilla deletion and random deletion, respectively. 

\end{blackText}

\subsubsection{Method}

\textbf{Deletion} replaces $t \times t$ size grid regions in an input image with a constant value (e.g., an average pixel value) iteratively until the entire grid is replaced. Replacing order is determined by their importance rank. Given an attribution map, every region is divided into grids of the same size $t \times t$. Then, each grid is ranked by the average of attributions in them. As grids in the input image are gradually substituted with a constant value, and the new input is fed to the network, the output score shall drop step by step. By observing the score drop curve, one can quantify the faithfulness of the attribution map against the model. 

Originally, removing the most relevant grid first (MoRF) is introduced in \citet{Samek2017deletion}. However, ~\citet{Ancona2017LeRF} found that the grid that makes the model score drop the most rapidly is not necessarily an important grid. They proposed to reverse the removing order - remove the least important grid first (LeRF). LeRF expects the model score to drop gradually at first, and steeper through iteration. It measures \textit{false negative} of the attribution map in that LeRF deletion gives a low score if the attribution map assigns low relevant scores to important features of the model.

\begin{table}[t]
\centering
\caption{Classification top-1 accuracy (\%) of the baseline (ViT-B/16-224) and models trained with mixed sample data augmentation strategies. }
\label{tab:performance_vit}
\begin{adjustbox}{width=0.28\textwidth}
    \begin{tabular}{lc}
        \noalign{\smallskip}\noalign{\smallskip}\hline\hline
        Methods & Top-1 accuracy (\%) \\
        \hline
        Baseline & 81.20 \\ 
        HTM &82.37  \\
        VTM &82.30  \\
        HTM + VTM &82.32  \\
        
        \hline
        \hline
    \end{tabular}
\end{adjustbox}
\end{table}

In our LeRF experiment, a slow decrease at early iteration steps in the model's score tells two different things in two different subjects. The first category is the interpretability of the attribution method. It suggests that the attribution method is \textit{faithful} to the model. Since removing the region where the attribution method pointed as redundant caused little change in the model score, it can be interpreted that the model takes little into account the region when making the decision. The second category is the occlusion robustness of the model. It suggests the model is \textit{robust to occlusions}. If removing small parts of the original input did not significantly affect the model, it suggests that the model is robust to occlusions. Reversely, the output score's steep drop at early steps and gradual drop at later steps can be interpreted as less faithful, or more sensitive. 
In other words, the deletion experiment shows the mixed effects of faithfulness and occlusion robustness. This is not a problem when the task is to compare the faithfulness of different attribution methods on the \emph{same} model. However, what we are doing is quantifying the attribution map faithfulness to \emph{different} models (different parameter sets trained from different augmentation schemes on the same architecture). 

Therefore, to analyze faithfulness under the reduced effect of occlusion robustness, we propose Inter-Model Deletion. 
Inter-Model Deletion introduces an additional procedure that selects the replaced grid in Random Order (RaO) in the deletion experiment as depicted in Figure \ref{fig:inter-del}b. We repeat five different random orders per image to alleviate the impact of randomness. The rest of the settings are the same as LeRF. RaO tells only about occlusion robustness because the grids do not contain any information about the relevance scores created by the attribution map. Then, we obtain a new plot that the effect of occlusion robustness is removed by subtracting RaO from LeRF (Figure \ref{fig:inter-del}c). Lastly, the area under the curve (AUC) is calculated for the final faithfulness score. 

\begin{blackText}
Specifically, Inter-Model Deletion is composed of three submodules: Grid Mapping Module, Sequence Module, and Deletion Module. 

\textbf{Grid Mapping module} maps the attribution map into grids size of ${t}$ x ${t}$. When an explanation (i.e., attribution map) is denoted by $E$, the output of the grid mapping module $E^{*}$ can be expressed in matrix form where each element of $m^{th}$ row and $n^{th}$ column represent the average of pixel values of the explanation in the corresponding position. This mapping is achieved by mapping function $\phi$ that transforms the grid matrix index $(m,n)$ into Cartesian coordinates $(u,v)$ within the two-dimensional space, which can be expressed $\phi: \mathcal{I} \rightarrow \mathbb{R}^2$ where $\phi(m, n) = (u, v) = (tn, tm)$ for a shared origin. 

\begin{equation}
\begin{aligned}
e_{m,n} = \frac{1}{t^{2}} \sum {E[u:u+t, v:v+t]} \\
\end{aligned}
\end{equation}

\noindent where \( E^{*} = [e_{m,n}] \) for \( 0 \leq m, n \leq T - 1 \) when the input size is ${N}$ x ${N}$ and total number of grids is $T^{2} = \frac{N}{t} \times \frac{N}{t}$.

\textbf{Sequence Module} determines the sequence order of the deletion. We use two strategies to set the sequence: LeRF and RaO. The output of the LeRF sequence module $\mathcal{S}_\text{LeRF}$ is the sorted index in ascending of the element values of $E^{*}$. In contrast, RaO sequence module $\mathcal{S}_\text{RaO}$ randomly generates the index. 
When indices of the matrix $E^{*}$ is defined by $\mathcal{I} = \{ (m,n) \mid 0 \leq m, n \leq T - 1 \}$,
we sort indices by their values to output the sequence of deletion order as

\begin{equation}
\begin{aligned}
\mathcal{S}_\text{LeRF} = \{ (m_1, n_1), (m_2, n_2), \ldots, (m_{T^2}, n_{T^2}) \}
\end{aligned}
\end{equation}

\noindent such that $e_{m_1, n_1} \leq e_{m_2, n_2} \leq \ldots \leq e_{m_{T^2}, n_{T^2}}$.
For RaO method, the deletion sequence $\mathcal{S}_\text{RaO}$ is generated randomly without referring to the explanation $E^{*}$ as

\begin{equation}
\begin{aligned}
\mathcal{S}_\text{RaO} = \pi(\{ (m,n) \mid 0 \leq m, n \leq T - 1 \})
\end{aligned}
\end{equation}

\noindent where $\pi$ denotes a random permutation of the original index set, providing a non-deterministic order for the deletion process. 

\textbf{Deletion Module} $\mathcal{G}$ iterates through the input image $x$ and the sequence module $\mathcal{S}$ to delete regions in $x$ that corresponds to the grid index. The output of the deletion module at the iteration step $i$ is given by

\begin{equation}
\begin{aligned}
x^{i} = \mathcal{G}(x^{i-1}, \mathcal{S}^{i})
\end{aligned}
\end{equation}

\noindent where $x^{i}$ denotes the perturbed input image at the iteration step $i$ step and thereby $x^{0} = x$. $\mathcal{S}^{i}$ is the grid index to be deleted from the input at $i^{th}$ step, which can be mapped to Cartesian coordinate on the input image space through $\phi$ so that $\phi(m_i, n_i) = (tu_i, tv_i)$. Then, the iterative opertation of deletion module can be written by

\begin{equation}
\begin{aligned}
x^{i-1}[u_i:u_i+t, v_i:v_i+t] = c. \\
\end{aligned}
\end{equation}

Let $x_\text{LeRF}^{i}$ be the deleted image after $i^{th}$ iteration on LeRF deletion and $x_\text{RaO}^{i}$ be the deleted image after $i^{th}$ iteration on RaO deletion. Then, the final inter-model deletion score can be obtained by the difference between the two deletion curves along with the deletion step $L$.

\begin{equation}
\begin{aligned}
\int \left[ f(x_\text{LeRF}^{i} - f(x_\text{RaO}^{i}) \right] dL.
\end{aligned}
\end{equation}

\end{blackText}

\subsubsection{Results}

To demonstrate the limitations of vanilla LeRF and the difference to Inter-Model Deletion, we show the Pearson correlation between RaO deletion and attribution map-based deletion (i.e., LeRF and Inter-Model Deletion) scores from each model in averaged over all probed attribution methods in Table \ref{tab:correlation}. The exact deletion values for each model against each attribution method are provided in Table \ref{tab:vanilla_deletion}. 
Table \ref{tab:correlation}(a) shows that the vanilla LeRF score is highly correlated with RaO, which suggests that the level of sensitivity against deletion, referred to as occlusion robustness in this paper, is reflected in the LeRF score. It is important to note that different models have different occlusion robustness and models trained with mixed sample data augmentation tend to be more robust against this kind of deletion \citep{Yun2019CutMix}. 
This raises the question of whether LeRF is suitable for assessing the attribution faithfulness of different models given that each model's sensitivity to deletion is highly reflected in LeRF scores. 
On the contrary, Table \ref{tab:correlation}(b) illustrates that Inter-Model Deletion effectively mitigates the influence of occlusion robustness in its deletion process, showing the decreased correlation with RaO scores on both CIFAR-10 and ImageNet.

The experimental result of Inter-Model Deletion is given in Table \ref{inter-deletion}. The grid size is set to $7 \times 7$ for ImageNet and $2 \times 2$ for CIFAR-10.
On both datasets, we observe that models trained with mixed sample data augmentation tend to degrade faithfulness. The baseline showed the best faithfulness on CIFAR-10 and the second-best on ImageNet. 
On CIFAR-10, Pixmix showed the second faithfulness score followed by RecursiveMix and SaliencyMix. Here, the average score drop of the second-best model (i.e., PixMix) over the best model (i.e., baseline) is statistically significant ($p < 0.001$). 
On ImageNet, RecursiveMix archives higher scores than Baseline on average, but the difference is not significant ($p = 0.227$) while the score improvement of Baseline over Mixup (third-best) is statistically significant ($p < 0.001$). In other words, the experimental results reveal false negatives of attribution maps created by models that are trained with mixed sample augmentation (i.e., the attribution method assigns low importance scores even though the feature is deemed important to the model).

\begin{blackText}
In Table \ref{tab:vanilla_deletion}, on the experiment result from ImageNet, we can see that Mixup (68.45) shows a higher vanilla LeRF score than CutMix (65.39) even though Mixup is more vulnerable to occlusions. On this, it is easily assumed that explanations from Mixup are more faithful than that of CutMix. Evaluation on Inter-Model deletion in Table \ref{inter-deletion} successfully captures this by evaluating higher scores on Mixup than CutMix. 
Meanwhile, Mixup shows a higher vanilla LeRF score (68.45) than the baseline (63.14). However, in comparison with the baseline, Mixup is more robust to occlusions. In this case, we cannot easily conclude that the explanation of Mixup is more faithful than that of the baseline as the high LeRF score might be the result of strong occlusion robustness. Evaluation on Inter-Model Deletion in Table \ref{inter-deletion} reveals that explanations of baseline are in fact more faithful.

\end{blackText}

\subsection{Analysis on Transformer-based Models}
This subsection presents the effect of mixed sample data augmentation on transformer-based models.
\paragraph{\textbf{Experimental Setup}} 
 We examined ViT-B/16-224 \citep{dosovitskiy2020vit} pre-trained on ImageNet-21k, fine-tuned on ImageNet-1k, and further fine-tuned with TokenMixup. TokenMixup includes Horizontal TokenMixup (HTM), Vertical TokenMixup (VTM), and a combination of both. We used official pre-trained weights for the baseline. For all models trained with TokenMixup, models are fine-tuned for 30 epochs with a batch size of 504, and SGD optimizer is used. The learning rate of HTM is scheduled with cosine annealing where the maximum and minimum learning rate is set to 0.015 and 0.0015, respectively. For VTM, the learning rate is scheduled with cosine annealing from 3e-2 to 3e-3. For more detailed training conditions, please refer to the original paper. The official classification performance of assessed models is given in Table \ref{tab:performance_vit}. 
 
\paragraph{\textbf{Results}} 

The experiment on localization is shown in Table \ref{localization_vit}. We found that the baseline scored the highest localization score and the combined usages of HTM and VTM give the lowest scores on both EnergyPG and EHR. A similar tendency is observed in the faithfulness experiment (Table \ref{inter-deletion_vit}) where the baseline with no mixed sample data augmentation scored the best.

\begin{table*}[t]
\centering
\renewcommand{\arraystretch}{1}
\caption{EnergyPG and EHR score of four models, trained with TokenMixup variants on ImageNet. The models are evaluated by four feature attribution methods, respectively. The base structure of the models is Vit-B/16-224. The best and second-best results are shown in bold, and underlined, respectively. The average of four different attribution methods is reported. The standard errors are also presented. }
\label{localization_vit}
\begin{tabular}{cc|cccc|c}
\noalign{\smallskip}\noalign{\smallskip}\hline\hline

Metric &  Methods & GradCAM & SmoothGrad & IntGrad & GBP & \textbf{Average} 
\\
\hline
\multirow{4}{*}{EnergyPG}
        &Baseline  & \textbf{0.501}\begin{scriptsize}$\pm$0.006\end{scriptsize} 
         & \textbf{0.465}\begin{scriptsize}$\pm$0.003\end{scriptsize} 
         & \textbf{0.398}\begin{scriptsize}$\pm$0.003\end{scriptsize} 
         & 0.363\begin{scriptsize}$\pm$0.003\end{scriptsize} 
         & \textbf{0.432}\begin{scriptsize}$\pm$0.002\end{scriptsize} 
          \\ 
        &HTM & 0.420\begin{scriptsize}$\pm$0.005\end{scriptsize} 
         & 0.455\begin{scriptsize}$\pm$0.003\end{scriptsize} 
         & \underline{0.391}\begin{scriptsize}$\pm$0.003\end{scriptsize} 
         & \textbf{0.366}\begin{scriptsize}$\pm$0.003\end{scriptsize} 
         & 0.408\begin{scriptsize}$\pm$0.002\end{scriptsize} 
         \\
        &VTM & \underline{0.460}\begin{scriptsize}$\pm$0.005\end{scriptsize} 
         & \underline{0.456}\begin{scriptsize}$\pm$0.003\end{scriptsize} 
         & 0.386\begin{scriptsize}$\pm$0.003\end{scriptsize} 
         & 0.363\begin{scriptsize}$\pm$0.003\end{scriptsize} 
         & \underline{0.416}\begin{scriptsize}$\pm$0.002\end{scriptsize} 
         \\ 
        &HTM + VTM & 0.415\begin{scriptsize}$\pm$0.005\end{scriptsize} 
         & 0.450\begin{scriptsize}$\pm$0.003\end{scriptsize} 
         & 0.388\begin{scriptsize}$\pm$0.003\end{scriptsize} 
         & \underline{0.365}\begin{scriptsize}$\pm$0.003\end{scriptsize} 
         & 0.405\begin{scriptsize}$\pm$0.002\end{scriptsize} 
         \\ 
\hline
\multirow{4}{*}{EHR}  
        &Baseline  & \textbf{0.331}\begin{scriptsize}$\pm$0.005\end{scriptsize} 
        & \textbf{0.222}\begin{scriptsize}$\pm$0.002\end{scriptsize}
        & \textbf{0.135}\begin{scriptsize}$\pm$0.002\end{scriptsize}
        & 0.105\begin{scriptsize}$\pm$0.002\end{scriptsize}
        & \textbf{0.198}\begin{scriptsize}$\pm$0.002\end{scriptsize} \\ 
        &HTM & 0.281\begin{scriptsize}$\pm$0.004\end{scriptsize}
        & 0.219\begin{scriptsize}$\pm$0.002\end{scriptsize}
        & 0.123\begin{scriptsize}$\pm$0.002\end{scriptsize}
        & \textbf{0.107}\begin{scriptsize}$\pm$0.002\end{scriptsize}
        & 0.183\begin{scriptsize}$\pm$0.002\end{scriptsize}\\
        &VTM & \underline{0.315}\begin{scriptsize}$\pm$0.004\end{scriptsize}
        & \underline{0.221}\begin{scriptsize}$\pm$0.002\end{scriptsize}
        & 0.118\begin{scriptsize}$\pm$0.002\end{scriptsize}
        & 0.105\begin{scriptsize}$\pm$0.002\end{scriptsize}
        & \underline{0.190}\begin{scriptsize}$\pm$0.002\end{scriptsize}
        \\ 
        &HTM + VTM & 0.281\begin{scriptsize}$\pm$0.004\end{scriptsize}
        & 0.214\begin{scriptsize}$\pm$0.002\end{scriptsize}
        & \underline{0.124}\begin{scriptsize}$\pm$0.002\end{scriptsize}
        & \underline{0.106}\begin{scriptsize}$\pm$0.002\end{scriptsize}
        & 0.181\begin{scriptsize}$\pm$0.002\end{scriptsize}
        \\ 

\hline
\hline
\end{tabular}
\end{table*}

\begin{table*}[t]
\centering
\renewcommand{\arraystretch}{1}
\caption{Inter-Model Deletion results on ViT-B/16-224. The models are evaluated by four feature attribution methods, respectively. Higher is better. The best and second-best results are shown in bold, and underlined, respectively. The average of four different attribution methods is reported. The standard errors are also presented.}
\label{inter-deletion_vit}
\begin{tabular}{cc|cccc|c}
\noalign{\smallskip}\noalign{\smallskip}\hline\hline
Dataset & Methods & GradCAM & SmoothGrad & IntGrad & GBP & \textbf{Average} 
\\
\hline
\multirow{4}{*}{ImageNet} 
        & Baseline & \textbf{33.73}\begin{scriptsize}$\pm$0.52\end{scriptsize}
         & 43.79\begin{scriptsize}$\pm$0.38\end{scriptsize}
         & \textbf{35.12}\begin{scriptsize}$\pm$0.41\end{scriptsize}
         & \underline{24.54}\begin{scriptsize}$\pm$0.45\end{scriptsize}
         & \textbf{34.30}\begin{scriptsize}$\pm$0.23\end{scriptsize}
          \\ 
        & HVM & 25.33\begin{scriptsize}$\pm$0.58\end{scriptsize}
          & \underline{44.57}\begin{scriptsize}$\pm$0.38\end{scriptsize}
          & \underline{34.11}\begin{scriptsize}$\pm$0.40\end{scriptsize}
          & \textbf{24.93}\begin{scriptsize}$\pm$0.44\end{scriptsize}
          & 32.24\begin{scriptsize}$\pm$0.24\end{scriptsize}
           \\
        & VTM & \underline{32.21}\begin{scriptsize}$\pm$0.52\end{scriptsize}
          & \textbf{44.85}\begin{scriptsize}$\pm$0.38\end{scriptsize}
          & 32.73\begin{scriptsize}$\pm$0.41\end{scriptsize}
          & 24.26\begin{scriptsize}$\pm$0.44\end{scriptsize}
          & \underline{33.51}\begin{scriptsize}$\pm$0.23\end{scriptsize}
           \\ 
        & HVM + VTM & 24.23\begin{scriptsize}$\pm$0.57\end{scriptsize}
          & 43.53\begin{scriptsize}$\pm$0.38\end{scriptsize}
          & 32.89\begin{scriptsize}$\pm$0.40\end{scriptsize}
          & 24.16\begin{scriptsize}$\pm$0.44\end{scriptsize}
          & 31.20\begin{scriptsize}$\pm$0.24\end{scriptsize}
           \\

\hline
\hline
\end{tabular}
\end{table*}

\begin{blackText}
\section{Effect of Label Mixing on Model Interpretability}
\subsection{Image Mixing and Label Mixing}
So far, we observed that models trained with various mixed sample augmentation strategies degrade the interpretability of attribution maps in terms of localization and faithfulness. 
Many algorithms of mixed sample data augmentation strategies \citep{zhang2017mixup,Yun2019CutMix,uddin2020saliencyMix,yang2022recursivemix} can be decomposed into two parts - image mixing and label mixing. In the image mixing part, the mixed image is generated by mixing the original image with the other randomly selected image, and in the label mixing part, the mixed label is usually set by the proportion of labels of the mixed images.

To better understand whether the decrease in interpretability is due to mixed images, mixed labels, or a combination of both, we compare the interpretability of models trained with dissected Mixup variants: Img-Mix-Only Mixup, and Label-Mix-Only Mixup. In Img-Mix-Only Mixup, the image is mixed in a Mixup-way but the label is determined to be a one-hot based on the mixed sample with the larger proportion. On the other hand, Label-Mix-Only Mixup is trained with non-mixed images and mixed labels. Likewise, images are determined by the bigger proportion of the mixed labels. Experimental results suggest that label mixing plays an important role in the degradation of interpretability.

 The experimental results are given in Table \ref{tab:onehotmixup}. We initially expected a lower Inter-Model Deletion score on Img-Mix-Only Mixup and Label-Mix-Only Mixup because of the reduced alignment of image-label pairs. However, we did not observe the tendency. Experimental results revealed higher scores for dissected Mixup variants where there is a larger discrepancy between the image and the corresponding label. Moreover, the results show that label mixing has a greater impact on faithfulness than image mixing. Here, The mixing step of each of the four models can be considered in a cascading way. Starting from the Baseline, adding image mixing gives Img-Only-Mixup, and adding label mixing to Img-Only-Mixup gives Mixup. In the same manner, starting from the Baseline, applying label mixing leads to Label-Mix-Only Mixup, and applying image mixing to Label-Mix-Only Mixup leads to Mixup. Thus, the effect of label mixing is represented by the gap between Baseline vs. Label-mix-only Mixup (14.10 on CIFAR and 6.47 on ImageNet) and Img-Mix-Only Mixup vs. Mixup (6.80 on CIFAR and 1.35 on ImageNet), while the effect of image mixing is represented by the gap between Baseline vs. Img-mix-only Mixup (8.58 on CIFAR and 1.02 on ImageNet) and Label-mix-only Mixup vs. Mixup (1.28  on CIFAR and -4.1 on ImageNet).

Table \ref{tab:onehotmixup_loc} shows localization scores of the baseline and Mixup variants on ResNet-50 trained on ImageNet. We found higher or compatible localization scores with Mixup variants where there is a larger gap between the image and the label. On EnergyPG, the effect of label mixing is greater than the effect of image mixing where the former is represented in the score gap between Baseline vs. Label-mix-only Mixup and Image-mix-only Mixup and Mixup and the latter is represented by the score gap between Baseline vs. Image-mix-only Mixup and Label-mix-only Mixup and Mixup. On EHR, we found that the effect of label mixing and image mixing is compatible.

\begin{table*}[t]
\centering
\color{black}
\renewcommand{\arraystretch}{1}
        \caption{\textcolor{black}{Inter-Model Deletion score of models trained with Mixup and its dissected variants. The best and second-best results are shown in bold, and underlined, respectively. The average of five different attribution methods is reported. The standard errors are also presented.}}
\label{tab:onehotmixup}
\begin{tabular}{cc|ccccc|c}
\noalign{\smallskip}\noalign{\smallskip}\hline\hline
Dataset & Methods & GradCAM & IBA & SmoothGrad & IntGrad & GBP & \textbf{Average} 
\\
\hline
\multirow{4}{*}{CIFAR-10} 
    &Baseline & \textbf{27.87}\begin{scriptsize}$\pm$0.47\end{scriptsize}
     & \textbf{20.17}\begin{scriptsize}$\pm$0.49\end{scriptsize}
     & \textbf{32.39}\begin{scriptsize}$\pm$0.44\end{scriptsize}
     & \textbf{32.00}\begin{scriptsize}$\pm$0.38\end{scriptsize}
     & \textbf{25.21}\begin{scriptsize}$\pm$0.35\end{scriptsize}
     & \textbf{27.53}\begin{scriptsize}$\pm$0.20\end{scriptsize}
      \\ 
    &Img-Mix-Only Mixup & \underline{17.45}\begin{scriptsize}$\pm$0.35\end{scriptsize}
     & \underline{16.13}\begin{scriptsize}$\pm$0.35\end{scriptsize}
     & 21.54\begin{scriptsize}$\pm$0.33\end{scriptsize}
     & 21.10\begin{scriptsize}$\pm$0.29\end{scriptsize}
     & \underline{18.54}\begin{scriptsize}$\pm$0.28\end{scriptsize}
     & \underline{18.95}\begin{scriptsize}$\pm$0.14\end{scriptsize}
     \\
    &Label-Mix-Only Mixup & 3.16\begin{scriptsize}$\pm$0.27\end{scriptsize}
     & 4.61\begin{scriptsize}$\pm$0.27\end{scriptsize}
     & \underline{22.01}\begin{scriptsize}$\pm$0.28\end{scriptsize}
     & \underline{21.14}\begin{scriptsize}$\pm$0.28\end{scriptsize}
     & 16.24\begin{scriptsize}$\pm$0.24\end{scriptsize}
     & 13.43\begin{scriptsize}$\pm$0.14\end{scriptsize}
     \\
    & Mixup & 7.42\begin{scriptsize}$\pm$0.30\end{scriptsize}
     & 7.41\begin{scriptsize}$\pm$0.27\end{scriptsize}
     & 18.96\begin{scriptsize}$\pm$0.30\end{scriptsize}
     & 17.34\begin{scriptsize}$\pm$0.26\end{scriptsize}
     & 9.60\begin{scriptsize}$\pm$0.21\end{scriptsize}
     & 12.15\begin{scriptsize}$\pm$0.13\end{scriptsize}
      \\ 

\hline

\multirow{4}{*}{ImageNet}
    &Baseline  & \underline{57.55}\begin{scriptsize}$\pm$0.33\end{scriptsize}
    & \textbf{58.75}\begin{scriptsize}$\pm$0.34\end{scriptsize}
    & 53.38\begin{scriptsize}$\pm$0.37\end{scriptsize}
    & \textbf{44.35}\begin{scriptsize}$\pm$0.42\end{scriptsize}
    & \textbf{46.60}\begin{scriptsize}$\pm$0.39\end{scriptsize}
    & \textbf{52.13}\begin{scriptsize}$\pm$0.18\end{scriptsize}
     \\
    &Img-Mix-Only Mixup & \textbf{58.39}\begin{scriptsize}$\pm$0.30\end{scriptsize}
     & 57.04\begin{scriptsize}$\pm$0.34\end{scriptsize}
     & \textbf{54.70}\begin{scriptsize}$\pm$0.34\end{scriptsize}
     & \underline{39.65}\begin{scriptsize}$\pm$0.42\end{scriptsize}
     & \underline{45.79}\begin{scriptsize}$\pm$0.39\end{scriptsize}
     & \underline{51.11}\begin{scriptsize}$\pm$0.18\end{scriptsize}
     \\
    &Label-Mix-Only Mixup & 50.58\begin{scriptsize}$\pm$0.27\end{scriptsize}
     & 52.59\begin{scriptsize}$\pm$0.29\end{scriptsize}
     & 49.39\begin{scriptsize}$\pm$0.30\end{scriptsize}
     & 39.15\begin{scriptsize}$\pm$0.37\end{scriptsize}
     & 36.59\begin{scriptsize}$\pm$0.33\end{scriptsize}
     & 45.66\begin{scriptsize}$\pm$0.15\end{scriptsize}
     \\
         &Mixup & 56.09\begin{scriptsize}$\pm$0.30\end{scriptsize}
     & \underline{58.27}\begin{scriptsize}$\pm$0.31\end{scriptsize}
     & \underline{54.50}\begin{scriptsize}$\pm$0.35\end{scriptsize}
     & 40.63\begin{scriptsize}$\pm$0.45\end{scriptsize}
     & 39.33\begin{scriptsize}$\pm$0.40\end{scriptsize}
     & 49.76\begin{scriptsize}$\pm$0.18\end{scriptsize}
      \\

\hline
\hline
\end{tabular}
\end{table*}


\begin{table*}[t]
\centering
\color{black}
\renewcommand{\arraystretch}{1}
\caption{\textcolor{black}{Localization scores of ResNet-50 trained on ImageNet with Mixup and its dissected variants. The best and second-best results are shown in bold, and underlined, respectively. The average of five different attribution methods is reported. The standard errors are also presented.}}
\label{tab:onehotmixup_loc}
\begin{tabular}{cc|ccccc|c}
\noalign{\smallskip}\noalign{\smallskip}\hline\hline

Metric & Method & GradCAM & IBA & SmoothGrad & IntGrad & GBP & \textbf{Average} 
\\
\hline
\multirow{4}{*}{EnergyPG}
         &Baseline & \textbf{0.549}\begin{scriptsize}$\pm$0.004\end{scriptsize} & \textbf{0.637}\begin{scriptsize}$\pm$0.004\end{scriptsize} & 
         0.494\begin{scriptsize}$\pm$0.003\end{scriptsize} & \textbf{0.465}\begin{scriptsize}$\pm$0.003\end{scriptsize} & 
         \textbf{0.651}\begin{scriptsize}$\pm$0.004\end{scriptsize} &
         \textbf{0.559}\begin{scriptsize}$\pm$0.002\end{scriptsize} 
         \\ 
        &Img-Mix-Only Mixup & 0.542\begin{scriptsize}$\pm$0.004\end{scriptsize} & 
        0.602\begin{scriptsize}$\pm$0.004\end{scriptsize} & 
        \textbf{0.512}\begin{scriptsize}$\pm$0.003\end{scriptsize} & 
        0.424\begin{scriptsize}$\pm$0.003\end{scriptsize} & 
        \underline{0.589}\begin{scriptsize}$\pm$0.004\end{scriptsize} & 
        \underline{0.534}\begin{scriptsize}$\pm$0.002\end{scriptsize} 
        \\
        &Label-Mix-Only Mixup & \underline{0.547}\begin{scriptsize}$\pm$0.004\end{scriptsize} 
         & 0.554\begin{scriptsize}$\pm$0.004\end{scriptsize} 
         & \underline{0.495}\begin{scriptsize}$\pm$0.003\end{scriptsize} 
         & \underline{0.446}\begin{scriptsize}$\pm$0.003\end{scriptsize} 
         & 0.487\begin{scriptsize}$\pm$0.004\end{scriptsize} 
         & 0.506\begin{scriptsize}$\pm$0.002\end{scriptsize} 
          \\ 
        
         &Mixup & 0.530\begin{scriptsize}$\pm$0.004\end{scriptsize} & 
         \underline{0.608}\begin{scriptsize}$\pm$0.004\end{scriptsize} & 0.485\begin{scriptsize}$\pm$0.003\end{scriptsize} & 0.420\begin{scriptsize}$\pm$0.003\end{scriptsize} & 0.403\begin{scriptsize}$\pm$0.004\end{scriptsize} &
         0.489\begin{scriptsize}$\pm$0.002\end{scriptsize}
        \\

\hline
\multirow{4}{*}{EHR}
        &Baseline & 0.484\begin{scriptsize}$\pm$0.003\end{scriptsize} 
        &\textbf{0.449}\begin{scriptsize}$\pm$0.003\end{scriptsize} 
        & 0.255\begin{scriptsize}$\pm$0.002\end{scriptsize} 
        & \underline{0.187}\begin{scriptsize}$\pm$0.002\end{scriptsize} 
        &\textbf{0.254}\begin{scriptsize}$\pm$0.002\end{scriptsize}
        &\textbf{0.326}\begin{scriptsize}$\pm$0.002\end{scriptsize}
         \\
        &Img-Mix-Only Mixup & \underline{0.493}\begin{scriptsize}$\pm$0.003\end{scriptsize}
        & 0.396\begin{scriptsize}$\pm$0.003\end{scriptsize}
        & \underline{0.267}\begin{scriptsize}$\pm$0.002\end{scriptsize}
        & 0.163\begin{scriptsize}$\pm$0.002\end{scriptsize}
        & \underline{0.223}\begin{scriptsize}$\pm$0.002\end{scriptsize}
        & \underline{0.308}\begin{scriptsize}$\pm$0.002\end{scriptsize}
        \\ 
        &Label-Mix-Only Mixup & \textbf{0.509}\begin{scriptsize}$\pm$0.004\end{scriptsize}
        & \underline{0.428}\begin{scriptsize}$\pm$0.003\end{scriptsize}
        & 0.264\begin{scriptsize}$\pm$0.002\end{scriptsize}
        & 0.181\begin{scriptsize}$\pm$0.002\end{scriptsize}
        & 0.154\begin{scriptsize}$\pm$0.002\end{scriptsize}
        & 0.307\begin{scriptsize}$\pm$0.002\end{scriptsize}
        \\ 
        &Mixup & 0.477\begin{scriptsize}$\pm$0.003\end{scriptsize} & 
        0.404\begin{scriptsize}$\pm$0.003\end{scriptsize} & 
        \textbf{0.272}\begin{scriptsize}$\pm$0.002\end{scriptsize} & 
        \textbf{0.272}\begin{scriptsize}$\pm$0.002\end{scriptsize} &
        0.114\begin{scriptsize}$\pm$0.002\end{scriptsize} &
         \underline{0.308}\begin{scriptsize}$\pm$0.002\end{scriptsize}\\

\hline
\hline
\end{tabular}
\end{table*}


\subsection{Label Smoothing}

In this subsection, we adopt label smoothing to further investigate the effect of mixing in a more generalized setting. They are known to improve the calibration and knowledge distillation performance and reduce the impact of noisy labels \citep{liang2024comprehensive, muller2019does, gou2021knowledge}. We compared the interpretability of models trained using various degrees of label smoothing where the labels are smoothed by a mixture of the original label and uniform distribution as in \citet{szegedy2016rethinking} and \citet{muller2019does}. The label in label smoothing is defined as $y_c^{LS} = y_c(1-\alpha) + \alpha/C$ where parameter $\alpha$ determines the amount of label smoothing and $C$ denotes the number of classes. $\alpha$ = 0 represents the model which is trained with one-hot label. The experiment is conducted by setting $\alpha$ in the range of [0, 0.1, 0.3, 0.5].

Table \ref{tab:labelsmoothing} shows Inter-Model Deletion scores of the model trained with various degrees of label smoothing. The baseline model is trained with one-hot labels (i.e., no label smoothing). 
Experimental results indicate that the baseline without label smoothing achieves the highest faithfulness across all attribution methods. Moreover, the bigger $\alpha$ leads to lower Inter-Model Deletion scores. In other words, a higher level of label smoothing introduces a bigger misalignment between the explanation and the prediction and results in a smaller Inter-Model Deletion score.

On localization evaluation given in Table \ref{tab:labelsmoothing_loc}, shows a similar tendency. The baseline with no label smoothing showed the best localization scores on both EnergyPG and EHR and the localization scores decreased with increasing degrees of label smoothing.


\begin{table*}[t]
\centering
\color{black}
\renewcommand{\arraystretch}{1}
        \caption{\textcolor{black}{Inter-Model Deletion score for baseline models trained with different degrees of label smoothing, where $\alpha$ determines the degree of smoothing. The best and second-best results are shown in bold, and underlined, respectively. The average of five different attribution methods is reported. The standard errors are also presented.}}
\label{tab:labelsmoothing}
\begin{tabular}{cc|ccccc|c}
\noalign{\smallskip}\noalign{\smallskip}\hline\hline
Dataset  & Methods & GradCAM & IBA & SmoothGrad & IntGrad & GBP & \textbf{Average} 
\\
\hline
\multirow{4}{*}{CIFAR-10} 
    &Baseline ($\alpha = 0$) & \textbf{27.87}\begin{scriptsize}$\pm$0.47\end{scriptsize}
     & \textbf{20.17}\begin{scriptsize}$\pm$0.49\end{scriptsize}
     & \textbf{32.39}\begin{scriptsize}$\pm$0.44\end{scriptsize}
     & \textbf{32.00}\begin{scriptsize}$\pm$0.38\end{scriptsize}
     & \textbf{25.21}\begin{scriptsize}$\pm$0.35\end{scriptsize}
     & \textbf{27.53}\begin{scriptsize}$\pm$0.20\end{scriptsize}
      \\ 
    &$+ \alpha = 0.1$ & \underline{16.91}\begin{scriptsize}$\pm$0.43\end{scriptsize}
     & \underline{13.97}\begin{scriptsize}$\pm$0.44\end{scriptsize}
     & \underline{29.03}\begin{scriptsize}$\pm$0.37\end{scriptsize}
     & \underline{29.35}\begin{scriptsize}$\pm$0.33\end{scriptsize}
     & \underline{22.84}\begin{scriptsize}$\pm$0.31\end{scriptsize}
     & \underline{22.42}\begin{scriptsize}$\pm$0.18\end{scriptsize}
     \\
    &$+ \alpha = 0.3$ & 6.76\begin{scriptsize}$\pm$0.34\end{scriptsize}
     & 10.66\begin{scriptsize}$\pm$0.33\end{scriptsize}
     & 25.21\begin{scriptsize}$\pm$0.28\end{scriptsize}
     & 24.68\begin{scriptsize}$\pm$0.24\end{scriptsize}
     & 19.02\begin{scriptsize}$\pm$0.23\end{scriptsize}
     & 17.27\begin{scriptsize}$\pm$0.15\end{scriptsize}
     \\
    &$+ \alpha = 0.5$ & 0.14\begin{scriptsize}$\pm$0.29\end{scriptsize}
     & 7.63\begin{scriptsize}$\pm$0.29\end{scriptsize}
     & 18.43\begin{scriptsize}$\pm$0.20\end{scriptsize}
     & 17.50\begin{scriptsize}$\pm$0.17\end{scriptsize}
     & 14.11\begin{scriptsize}$\pm$0.17\end{scriptsize}
     & 11.56\begin{scriptsize}$\pm$0.12\end{scriptsize}
     \\

\hline

\multirow{4}{*}{ImageNet} 
    &Baseline ($\alpha = 0$) & \underline{57.55}\begin{scriptsize}$\pm$0.33\end{scriptsize}
    & \textbf{58.75}\begin{scriptsize}$\pm$0.34\end{scriptsize}
    & \textbf{53.38}\begin{scriptsize}$\pm$0.37\end{scriptsize}
    & \textbf{44.35}\begin{scriptsize}$\pm$0.42\end{scriptsize}
    & \textbf{46.60}\begin{scriptsize}$\pm$0.39\end{scriptsize}
    & \textbf{52.13}\begin{scriptsize}$\pm$0.18\end{scriptsize}
     \\
    &$+ \alpha = 0.1$ & \textbf{56.29}\begin{scriptsize}$\pm$0.32\end{scriptsize}
     & \underline{57.66}\begin{scriptsize}$\pm$0.30\end{scriptsize}
     & \underline{53.10}\begin{scriptsize}$\pm$0.34\end{scriptsize}
     & \underline{41.32}\begin{scriptsize}$\pm$0.41\end{scriptsize}
     & \underline{39.43}\begin{scriptsize}$\pm$0.38\end{scriptsize}
     & \underline{49.56}\begin{scriptsize}$\pm$0.18\end{scriptsize}
     \\
    &$+ \alpha = 0.3$ & 50.18\begin{scriptsize}$\pm$0.26\end{scriptsize}
     & 51.87\begin{scriptsize}$\pm$0.27\end{scriptsize}
     & 48.21\begin{scriptsize}$\pm$0.28\end{scriptsize}
     & 37.26\begin{scriptsize}$\pm$0.35\end{scriptsize}
     & 32.16\begin{scriptsize}$\pm$0.33\end{scriptsize}
     & 43.94\begin{scriptsize}$\pm$0.16\end{scriptsize}
     \\
    &$+ \alpha = 0.5$ & 45.37\begin{scriptsize}$\pm$0.32\end{scriptsize}
     & 46.36\begin{scriptsize}$\pm$0.34\end{scriptsize}
     & 43.83\begin{scriptsize}$\pm$0.29\end{scriptsize}
     & 33.85\begin{scriptsize}$\pm$0.38\end{scriptsize}
     & 27.83\begin{scriptsize}$\pm$0.41\end{scriptsize}
     & 39.45\begin{scriptsize}$\pm$0.17\end{scriptsize}
      \\

\hline
\hline
\end{tabular}
\end{table*}

\begin{table*}[t]
\centering
\color{black}
\renewcommand{\arraystretch}{1}
\caption{\textcolor{black}{Localization scores for baseline (ResNet-50) models trained on ImageNet with different degrees of label smoothing, where $\alpha$ determines the degree of smoothing. The best and second-best results are shown in bold, and underlined, respectively. The average of five different attribution methods is reported. The standard errors are also presented.}}
\label{tab:labelsmoothing_loc}
\begin{tabular}{cc|ccccc|c}
\noalign{\smallskip}\noalign{\smallskip}\hline\hline

Metric & Method & GradCAM & IBA & SmoothGrad & IntGrad & GBP & \textbf{Average} 
\\
\hline
\multirow{4}{*}{EnergyPG}
         &Baseline & \underline{0.549}\begin{scriptsize}$\pm$0.004\end{scriptsize} & \textbf{0.637}\begin{scriptsize}$\pm$0.004\end{scriptsize} & 
         \underline{0.494}\begin{scriptsize}$\pm$0.003\end{scriptsize} & 
         \textbf{0.465}\begin{scriptsize}$\pm$0.003\end{scriptsize} & 
         \textbf{0.651}\begin{scriptsize}$\pm$0.004\end{scriptsize} &
         \textbf{0.559}\begin{scriptsize}$\pm$0.002\end{scriptsize} 
         \\ 
        &$+ \alpha = 0.1$ & \textbf{0.555}\begin{scriptsize}$\pm$0.004\end{scriptsize} & 
        0.594\begin{scriptsize}$\pm$0.004\end{scriptsize} & 
        0.489\begin{scriptsize}$\pm$0.003\end{scriptsize} & 
        0.436\begin{scriptsize}$\pm$0.003\end{scriptsize} & 
        \underline{0.492}\begin{scriptsize}$\pm$0.004\end{scriptsize} & 
        \underline{0.513}\begin{scriptsize}$\pm$0.002\end{scriptsize}  
        \\ 
        &$+ \alpha = 0.3$ & 0.547\begin{scriptsize}$\pm$0.004\end{scriptsize} & 
        \underline{0.596}\begin{scriptsize}$\pm$0.004\end{scriptsize} & 
        \textbf{0.501}\begin{scriptsize}$\pm$0.003\end{scriptsize} & 
        \underline{0.441}\begin{scriptsize}$\pm$0.003\end{scriptsize} & 
        0.442\begin{scriptsize}$\pm$0.004\end{scriptsize} & 
        0.505\begin{scriptsize}$\pm$0.002\end{scriptsize} 
          \\ 
        &$+ \alpha = 0.5$ & 0.527\begin{scriptsize}$\pm$0.004\end{scriptsize} & 
        0.534\begin{scriptsize}$\pm$0.004\end{scriptsize} & 
        0.488\begin{scriptsize}$\pm$0.003\end{scriptsize} & 
        \underline{0.441}\begin{scriptsize}$\pm$0.003\end{scriptsize} & 
        0.368\begin{scriptsize}$\pm$0.004\end{scriptsize} & 
        0.472\begin{scriptsize}$\pm$0.002\end{scriptsize} 
          \\

\hline
\multirow{4}{*}{EHR}
        &Baseline & 0.484\begin{scriptsize}$\pm$0.003\end{scriptsize} 
        &\textbf{0.449}\begin{scriptsize}$\pm$0.003\end{scriptsize} 
        & 0.255\begin{scriptsize}$\pm$0.002\end{scriptsize} 
        & \textbf{0.187}\begin{scriptsize}$\pm$0.002\end{scriptsize} 
        &\textbf{0.254}\begin{scriptsize}$\pm$0.002\end{scriptsize}
        &\textbf{0.326}\begin{scriptsize}$\pm$0.002\end{scriptsize}
        \\
        &$+ \alpha = 0.1$  & \underline{0.499}\begin{scriptsize}$\pm$0.003\end{scriptsize}
        & 0.446\begin{scriptsize}$\pm$0.003\end{scriptsize}
        & \underline{0.263}\begin{scriptsize}$\pm$0.002\end{scriptsize}
        & 0.172\begin{scriptsize}$\pm$0.002\end{scriptsize}
        & \underline{0.156}\begin{scriptsize}$\pm$0.002\end{scriptsize}
         & \underline{0.307}\begin{scriptsize}$\pm$0.002\end{scriptsize}
        \\ 
        &$+ \alpha = 0.3$ & \textbf{0.502}\begin{scriptsize}$\pm$0.003\end{scriptsize}
        & \textbf{0.449}\begin{scriptsize}$\pm$0.003\end{scriptsize}
        & \textbf{0.266}\begin{scriptsize}$\pm$0.002\end{scriptsize}
        & \underline{0.174}\begin{scriptsize}$\pm$0.002\end{scriptsize}
         & 0.121\begin{scriptsize}$\pm$0.002\end{scriptsize}
        & 0.302\begin{scriptsize}$\pm$0.002\end{scriptsize}
        \\ 
        &$+ \alpha = 0.5$ & 0.482\begin{scriptsize}$\pm$0.004\end{scriptsize}
        & 0.419\begin{scriptsize}$\pm$0.003\end{scriptsize}
        & 0.262\begin{scriptsize}$\pm$0.002\end{scriptsize}
        & 0.172\begin{scriptsize}$\pm$0.002\end{scriptsize}
        & 0.087\begin{scriptsize}$\pm$0.002\end{scriptsize}
        & 0.285\begin{scriptsize}$\pm$0.002\end{scriptsize}
        \\ 

\hline
\hline
\end{tabular}
\end{table*}

To sum up, in Inter-Model Deletion experiments, we found that the effect of label mixing surpasses the effect of image mixing in the degradation of faithfulness. In the localization experiment, a similar tendency is observed on EnergyPG. However, on EHR, image mixing and label mixing have a similar degree of effect. This may be because the original score difference between the baseline and Mixup is relatively smaller in EHR than in EnergyPG, and thus, falls short of the discrimination ability between Mixup variants. 
Overall, the experimental results suggest that both input and label mixing may reduce interpretability, with labels playing a larger role. We further verified the effect of mixed labels by analyzing the baseline models trained with different degrees of label smoothing. We found that a higher degree of label smoothing introduces more degraded interpretability. 

This observation opens the door for a mixed sample data augmentation strategy that only mixes images, leaving labels one-hot in order to achieve compatible interpretability with the non-mixed counterparts. For mission-critical tasks where interpretability is as important as performance, mixed sample data augmentation variants with one-hot labels might be utilized as an alternative. 
For example, the medical domain is where providing an explanation of the model’s decision-making process is important \citep{wang2023pstcnn, singh2022think}. 
At the same time, it is one of the fields where large-scale data is hard to obtain \citep{chlap2021medicalReview}. DNNs are data-deprived \citep{mahajan2018exploring}, and data shortage can be alleviated by data augmentation. Therefore, some tasks in the medical domain where the interpretability in terms of attribution method is important may adopt mixed sample data augmentation with ont-hot labels to enjoy the benefits of mixed sample data augmentation with a moderating influence on interpretability.
\end{blackText}

\begin{blackText}

\section{Discussion}
\subsection{Maximizing Faithfulness of Explanations for Training}

In this section, we apply Inter-Model Deletion as an additional training objective to fine-tune models trained with mixed sample data augmentation strategies to enhance the overall interpretability of the models. This approach is in line with guided attention strategies, which are adopted in many tasks including classification \citep{dong2024ocie} and segmentation \citep{zhang2023attention, sinha2020multi, li2018tell}.

We define additional loss using inter-model deletion score $\mathcal{L}_{i}$ as the difference between probabilities between deleted inputs of two deleting strategies - LeRF and RaO.

\begin{equation}
\begin{aligned}
\mathcal{L}_{i}  = f(x_\text{LeRF}^{i}) - f(x_\text{RaO}^{i})
\end{aligned}
\end{equation}

\noindent where $f$ is the network, $x$ is the input, $x_\text{LeRF}^{i}$ and $x_\text{RaO}^{i}$ represents the deleted input at $i^{th}$ deletion step on LeRF and RaO strategy, respectively.  

We fine-tuned the pre-trained Mixup and CutMix models on MobileNetV2 on CIFAR-10 using the Deletion-based objective for 10 epochs. We use Grad-CAM to obtain attribution maps in training steps. The training results are presented in Table \ref{tab:inter-model-train}. As expected, the faithfulness of Grad-CAM on models trained with additional training objectives increased. Moreover, the overall faithfulness of other attribution methods also increases even when the model is trained to meet the Grad-CAM objective. This suggests that the overall explainability can be improved just by running a few additional epochs with inter-model deletion constraints. \\

\begin{table*}[t]
\centering
\color{black}
\renewcommand{\arraystretch}{1}
\caption{ \textcolor{black}{Faithfulness comparison on Inter-Model Deletion on models trained with and without faithfulness maximization objective. }}
\label{tab:inter-model-train}

\begin{tabular}{c|ccccc|c}
\noalign{\smallskip}\noalign{\smallskip}\hline\hline
Methods & GradCAM & IBA & SmoothGrad & IntGrad & GBP & \textbf{Average} 
\\
\hline

 Mixup & 7.42\begin{scriptsize}$\pm$0.30\end{scriptsize}
& 7.41\begin{scriptsize}$\pm$0.27\end{scriptsize}
& 18.96\begin{scriptsize}$\pm$0.30\end{scriptsize}
& 17.34\begin{scriptsize}$\pm$0.26\end{scriptsize}
& 9.60\begin{scriptsize}$\pm$0.21\end{scriptsize}
& 12.15\begin{scriptsize}$\pm$0.13\end{scriptsize}
\\ 
 
Faithful Mixup & \textbf{14.55}\begin{scriptsize}$\pm$0.50\end{scriptsize}
 & \textbf{12.54}\begin{scriptsize}$\pm$0.45\end{scriptsize}
 & \textbf{24.75}\begin{scriptsize}$\pm$0.44\end{scriptsize}
 & \textbf{24.34}\begin{scriptsize}$\pm$0.40\end{scriptsize}
 & \textbf{15.19}\begin{scriptsize}$\pm$0.32\end{scriptsize}
 & \textbf{18.27}\begin{scriptsize}$\pm$0.20\end{scriptsize}
 \\
\hline 

CutMix & 14.32\begin{scriptsize}$\pm$0.36\end{scriptsize}
& 9.47\begin{scriptsize}$\pm$0.36\end{scriptsize}
& 18.89\begin{scriptsize}$\pm$0.34\end{scriptsize}
& 21.02\begin{scriptsize}$\pm$0.30\end{scriptsize}
& 14.73\begin{scriptsize}$\pm$0.27\end{scriptsize}
& 15.69\begin{scriptsize}$\pm$0.15\end{scriptsize}
\\

Faithful CutMix & \textbf{19.39}\begin{scriptsize}$\pm$0.38\end{scriptsize}
 & \textbf{14.52}\begin{scriptsize}$\pm$0.38\end{scriptsize}
 & \textbf{20.76}\begin{scriptsize}$\pm$0.38\end{scriptsize}
 & \textbf{23.87}\begin{scriptsize}$\pm$0.32\end{scriptsize}
 & \textbf{17.14}\begin{scriptsize}$\pm$0.30\end{scriptsize}
 & \textbf{19.14}\begin{scriptsize}$\pm$0.16\end{scriptsize}
 \\

\hline
\hline
\end{tabular}
\end{table*}

\end{blackText}

\subsection{Limitation}

\textbf{Subject of evaluation.} In this paper, we conducted experiments only on vision data. The analysis of textual, audio, and multimodal data is an important subject for future research. 
We analyzed the interpretability of recent mixed-sample data augmentation strategies on various attribution methods. However, we did not conduct a survey-level study. There might be a sweet spot of combinations of datasets, mixed sample augmentation strategies, and attribution methods.

\textbf{Definining interpretability.} This study focuses on intrinsic (automatic) interpretability measures, which enable fair and objective evaluation. However, extrinsic evaluation, such as conducting surveys through Amazon Mechanical Turk to obtain human opinions on provided explanations or evaluating models that enable human intervention \citep{koh2020concept, taesiri2022humanaiteam}, will also be an important line of interpretability assessment \citep{Kim2022hive,Nguyen2021effectiveness} because one of the ultimate goals of explainable AI is to assist human users, e.g., model debugging or bias detecting for researchers, and trust building for common users. We examined interpretability when an explanation is given by an attribution map, but other definitions of interpretability, such as concept-based explanations \citep{bau2017dissection,kim2018interpretability} are also important aspects of interpretability. Therefore, it would be meaningful future work to investigate how mixed sample augmentation affects model interpretability from these perspectives.

\textbf{Theoretical study.} This work presents experimental findings regarding the interpretability of models trained with mixed sample data augmentation strategies. Furthermore, we found label smoothing (mixing) plays an important role. Theoretical study about the fundamental reason is going to be an interesting future research.

\section{Conclusion}

In this paper, we analyzed the unexplored side of mixed sample data augmentation methodologies: their effect on interpretability. More specifically, we examined how attribution maps are affected when models are trained with mixed sample data augmentation. We evaluated each model on localization and faithfulness. Throughout the experiments, we observed the degradation of interpretability in models trained with mixed sample data augmentation strategies. It suggests that attribution map errors (i.e., false positive, and false negative) have increased when models are trained with mixed sample augmentation. We found that label smoothing plays a significant role in degraded interpretability. 

However, our experimental results do \emph{not} suggest that researchers should stop using mixed sample data augmentation strategies. Rather, our work encourages the community to look at aspects that were overlooked before - their effect on interpretability. This is especially important for some applications where interpretation is critical. 
We highlight that
the community needs powerful data augmentation strategies that achieve high performance without having a negative influence on interpretability. 

\section*{Acknowledgements}
This work was partly supported by Institute of Information and Communications Technology Planning and Evaluation (IITP)-ITRC (Information Technology Research Center) grant funded by the Korea government (MSIT) (IITP-2025-RS-2023-00258649, 20\%), Institute of Information and Communications Technology Planning and Evaluation (IITP) grant funded by the Korea government (MSIT) (No. RS-2022-II220078: Explainable Logical Reasoning for Medical Knowledge Generation 20\%, No. RS-2024-00509257: Global AI Frontier Lab 20\%, No. RS-2022-00155911: Artificial Intelligence Convergence Innovation Human Resources Development (Kyung Hee University) 10\%, No. RS-2021-II212068: Artificial Intelligence Innovation Hub 10\%). This work was supported in part by the National Research Foundation of Korea (NRF) grant funded by the Korea government (MSIT) (No. RS-2024-00334321, 20\%). Dr. Seong Tae Kim is a corresponding author.

\bibliographystyle{elsarticle-harv}

\end{document}